\documentclass[sigconf]{acmart}
\AtBeginDocument{%
  }

\setcopyright{acmlicensed}
\copyrightyear{2018}
\acmYear{2018}
\acmDOI{XXXXXXX.XXXXXXX}
\acmConference[Conference acronym 'XX]{Make sure to enter the correct
  conference title from your rights confirmation email}{June 03--05,
  2018}{Woodstock, NY}
\acmISBN{978-1-4503-XXXX-X/2018/06}

\usepackage{algorithm}
\usepackage{amsmath}
\usepackage{enumitem}
\usepackage{booktabs}
\usepackage{multirow}
\usepackage{subfigure}
\usepackage{threeparttable}
\usepackage[table]{xcolor}
\usepackage[utf8]{inputenc}
\usepackage[noend]{algpseudocode}
\usepackage{diagbox}
\usepackage{tikz}
\usetikzlibrary{arrows.meta,positioning}

\usepackage{xspace}
\newcommand{\ourmethod}{CAPMix\xspace}

\newcommand{\mypara}[1]{{\vspace{0.1mm}\noindent\textbf{#1}}}

\begin{document}

\title{\ourmethod: Robust KPI Anomaly Detection for AIOps in Noisy and Dynamic Environments}







%


\author{Xudong Mou\textsuperscript{1}, 
Rui Wang\textsuperscript{1}, 
Tiejun Wang\textsuperscript{1},
Zexin Wu\textsuperscript{1},
Fangda Guo\textsuperscript{2},
Jie Sun\textsuperscript{1},  
Shiru Chen\textsuperscript{3}, 
Penghao Zhang\textsuperscript{4},
Tiezi Zhang\textsuperscript{4},
Tianyu Wo\textsuperscript{1}, 
Hao Peng\textsuperscript{1}, 
Chunming Hu\textsuperscript{1},
Xudong Liu\textsuperscript{1}, 
Renyu Yang\textsuperscript{1\dag}
}

\thanks{Corresponding Author: Renyu Yang (renyuyang@buaa.edu.cn)}

\affiliation{%
$^1$Beihang University \quad $^2$ Chinese Academy of Sciences\quad
$^3$Inspur Inc. \quad $^4$ Kuaishou Technology \country{}
}

\renewcommand{\shortauthors}{Mou et al.}

\begin{abstract}
Time-series anomaly detection is crucial in AIOps for maintaining large-scale service reliability.
In production, streams of Key Performance Indicators (KPI) are high-dimensional, non-stationary, and affected by noise, deployment changes, and latent anomalies, making real failures hard to distinguish from benign variation. Most existing methods assume either normality (learning from “normal” history) or rely on injected anomalies for training. Yet injected patterns often misalign with real failure modes, skewing decision boundaries -- aka.  \textit{Anomaly Shift}.
We propose \ourmethod, a controllable anomaly augmentation framework with prior-guided injection for realistic temporal behaviors. \ourmethod combines label revision and dual-space mixup to enhance robustness under contaminated and mixed data. \ourmethod consistently outperforms state-of-the-art methods on  public AIOps and time-series benchmarks. It has been deployed in Kuaishou's large-scale production system, reducing false alarms and improving monitoring reliability. A real-world dataset is also released to enrich the research on robust KPI anomaly detection. 
\end{abstract}



\received{20 February 2007}
\received[revised]{12 March 2009}
\received[accepted]{5 June 2009}

\maketitle

\section{Introduction}

Large-scale AI models have made high-performance computing clusters -- usually with over 10,000 GPUs -- mission-critical for services like real-time video recommendation and e-commerce. As these systems grow more complex, anomalies become more likely and more damaging, with even transient failures causing major revenue loss and disruption. To ensure reliability, AIOps platforms continuously monitor massive streams of Key Performance Indicators (KPIs), which form high-dimensional, time-varying series~\cite{zhao2021identifying, sun2023efficient, yu2024pre}. These sequences exhibit strong temporal dependencies, non-stationarity, and cross-metric correlations. 

In one of our production environments, we observed a task-level failure rate of 19.3\%. Interpretation of key performance indicator (KPI) signals is further confounded by measurement noise and by benign operations such as canary deployment and auto-scaling.
These factors introduce transient fluctuations and irregular patterns that depart from true anomalies, causing detectors to miss temporal structure, raise excessive false alarms, and create alert fatigue. This wastes computational resources and delays incident response and development. Hence, it is highly imperative for real-world KPI Anomaly Detection (KAD) to learn robust anomaly representations under noisy, evolving, weakly distinguishable temporal patterns.

\begin{figure}[t]
    \centering
    \includegraphics[width=0.98\columnwidth]{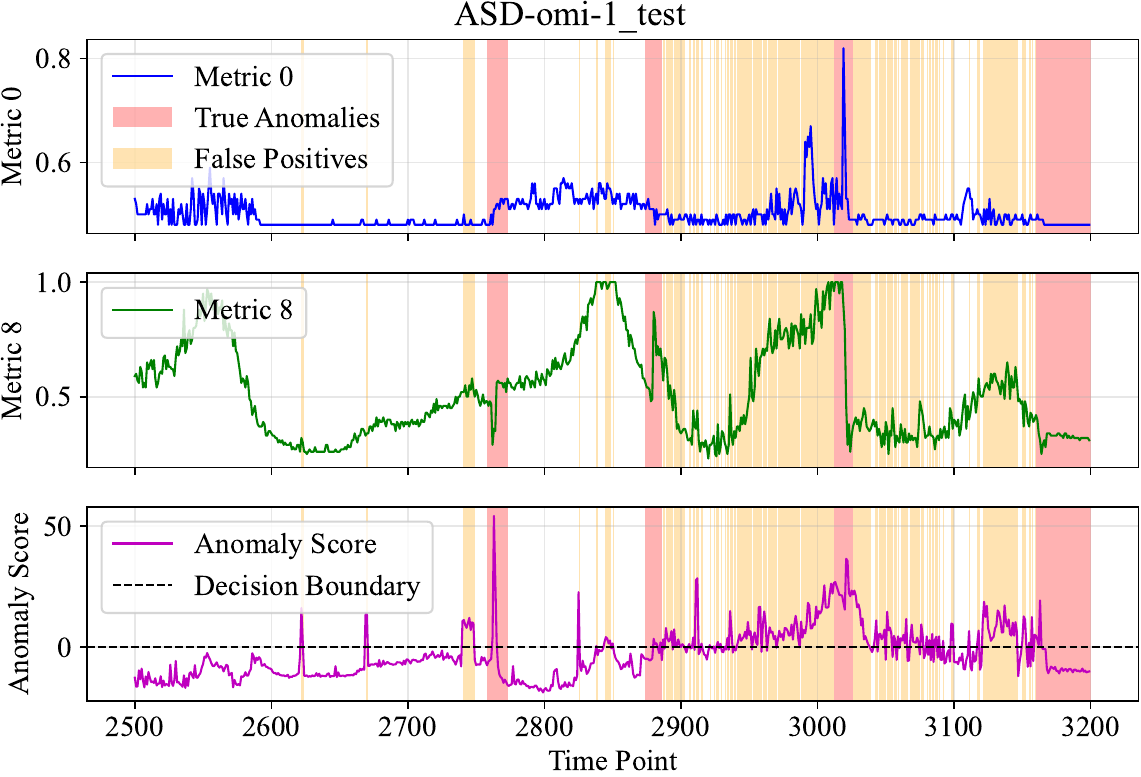}
    \vspace{-0.6em}
    \caption{False alarms of a normality-assumption detector OC-SVM on a representative KPI stream. The top panels show raw KPI sequences, and the bottom panel shows the OC-SVM anomaly score. Benign system fluctuations are repeatedly flagged as anomalies (orange), obscuring true anomalies (red) and causing an excessive false-positive rate.}
    \label{motivation}
    \vspace{-1em}
\end{figure}

Existing KPI anomaly detection methods usually rely on two implicit assumptions:

\textit{Normality Assumption (NA).} Many approaches assume that the training data is clean and mostly normal samples~\cite{malhotra2016lstm, ruff2018deep, yu2024pre, sun2023efficient}. Detectors learn a compact description of normal KPI dynamics and flag deviations as anomalies. For instance, reconstruction-based methods such as KAD-Disformer~\cite{yu2024pre} reconstruct the expected KPI curve and use the reconstruction error as the anomaly score. Other deviation-based frameworks, such as ART~\cite{sun2024art}, also model normal behavior and flag large deviations as anomalies.
However, production KPI streams are often noisy and may contain latent anomalies, violating NA and making the learned boundary overly sensitive and less robust to stochastic variations. As shown in Fig.~\ref{motivation}, a typical NA-based detector like OC-SVM can raise frequent false alarms on benign fluctuations; with a 5-minute sampling interval, these can persist all day. This limitation motivates incorporating explicit anomaly knowledge to better guide decision boundary learning.

\textit{Anomaly assumption (AA).} Some methods introduce anomaly patterns for supervision~\cite{hendrycks2018deep, li2021cutpaste, carmona2021neural, jeong2023anomalybert}.
Outlier Exposure (OE)~\cite{hendrycks2018deep} uses auxiliary outlier datasets as negative samples.
NCAD~\cite{carmona2021neural} and AnomalyBert~\cite{jeong2023anomalybert} inject random point- or context-level corruptions to mimic anomalies.
CutAddPaste~\cite{wang2024cutaddpaste} controllably injects five types of time-series anomalies.
Overall, AA effectively learns discriminative boundaries under contaminated data but requires sufficiently rich and well-calibrated anomaly patterns.
In practice, AA methods suffer from \textit{Anomaly Shift}, 
where injected anomaly patterns misalign with real failure modes.
It is due to either \textit{low distinctiveness} -- injected anomalies are insufficiently differentiated from normal operational patterns, 
resulting in missed detections -- or \textit{high deviation} where injected anomalies deviate excessively from realistic behaviors
thereby distorting the learned decision boundary.  
In the absence of ground-truth supervision, systematically calibrating both the injection intensity and the anomaly pattern space to adequately span the heterogeneous spectrum of industrial failure modes remains a non-trivial task. This challenge underscores the necessity for a controllable and robust anomaly generation framework that can better align synthetic anomalous patterns with real-world temporal dynamics.

We propose \ourmethod, a robust, controllable anomaly augmentation framework for KPI anomaly detection in contaminated, dynamic AIOps environments. \ourmethod uses prior-guided augmentation to align synthetic anomalies with real system behavior, injecting five types of realistic time-series anomalies in a targeted, controllable way. To mitigate \textit{Anomaly Shift}, we train a Temporal Convolutional Network (TCN) with two simple insights: Label Revision (LR) soft-labels normal-biased synthetic anomalies, and Dual-Space (DS) Mixup regularizes anomalies in both raw and latent spaces to better match real failures, together promoting smoother, more reliable decision boundaries between true failures and benign fluctuations. We evaluate \ourmethod on four public AIOps datasets, and four cross-domain time-series benchmarks indicate its potential to generalize from KPI streams to broader time-series anomaly detection tasks.
\ourmethod has been integrated and deployed as a part of KAIOps~\cite{wang2025kaiops}, the anomaly detector in Kuaishou's large-scale AI clusters. We also present case studies on critical failure modes and benign scaling-induced fluctuations, to demonstrate practical benefits in reducing alert fatigue and accelerating incident response.

The main contributions are as follows:
\begin{itemize}[leftmargin=1.2em]
\item A controllable framework that incorporates industrial prior knowledge into anomaly augmentation, bridging the gap between simulated and real-world failure patterns.
\item Dual-space mixup and label revision mechanisms to address the \textit{Anomaly Shift} problem, through learning robust decision boundaries under highly contaminated training data.
\item Industrial case studies to showcase improved detection accuracy and reduced false alarm rates in production environments.
\item Releasing an evaluation dataset with 16 days of production KPI streams, featuring real-world noise and deployment-induced dynamics.
\end{itemize}

Source code is available at \texttt{\url{https://github.com/alsike22/CAPMix}}.

\section{Related Work}
\label{sec:relatedwork}
KAD in AIOps builds on advances in time-series anomaly detection for software reliability engineering. We briefly review the most relevant work: (i) normality-assumption methods~\cite{pang2021deep}, (ii) anomaly-assumption/injection methods, and (iii) sample-mixing strategies for robustness. Table~\ref{relate_work} summarizes representative approaches, including KAD methods and general time-series anomaly detectors validated on KPI datasets, and highlights the gaps our method addresses.

\mypara{Normality assumption based methods.}
Normality-assumption (NA) methods assume most training windows are normal and learn a compact \textit{normality} model via GANs~\cite{schlegl2017unsupervised,xia2022gan}, autoencoders / reconstruction~\cite{malhotra2016lstm, feng2024sensitivehue, wucatch}, one-class objectives~\cite{ruff2018deep,xu2024calibrated,qiao2021efficient,wang2026moc}, or clustering~\cite{zong2018deep,zhou2024label}. Windows that deviate from this normal manifold are flagged as anomalous. In AIOps KPI anomaly detection, NA-style detectors are popular due to scarce labels. KAD-Disformer~\cite{yu2024pre}, for example, uses a pre-trained disentangled Transformer to reconstruct or estimate expected KPI patterns and treats reconstruction error as the anomaly signal. ART~\cite{sun2024art} similarly models normal microservice behavior within an incident-management framework and flags large deviations as abnormal. To broaden coverage, some methods combine multiple priors or signals: \cite{sohn2020learning} uses a two-stage procedure based on dataset-level representations, while COCA~\cite{wang2023deep} and RoCA~\cite{mou2025roca} fuse objectives to suppress features irrelevant to anomaly detection. However, NA methods can be over-sensitive to benign KPI turbulence (e.g., workload-driven oscillations), and the lack of explicit anomaly knowledge often forces models to relearn failure signatures implicitly.

\mypara{Anomaly assumption based methods.}
Rather than modeling only normality, anomaly-assumption methods use synthetic or real abnormal samples to directly shape the decision boundary. Outlier Exposure (OE)~\cite{hendrycks2018deep} trains with auxiliary out-of-distribution data, and Deep SAD~\cite{ruff2019deep} (extending Deep SVDD~\cite{ruff2018deep}) uses labeled anomalies to separate normal and abnormal representations. In CV, CutPaste~\cite{li2021cutpaste} simulates defects via cut-and-paste augmentation, and \cite{ruff2020rethinking} shows that even limited OE data can be effective.
For time series, NCAD~\cite{carmona2021neural} injects contextual and point anomalies to learn a classifier, but does not explicitly target richer structured anomalies. AnomalyBert~\cite{jeong2023anomalybert} perturbs points and subsequences to approximate both point- and pattern-wise anomalies. Generative methods such as GenIAS~\cite{darban2025genias} produce abnormal patches while discouraging normal-like generations. RedLamp~\cite{obata2025robust} applies diverse augmentations and label refurbishment to improve robustness under false anomalies/contamination, focusing on false-positive reduction. Our \ourmethod complements these works by (i) revising label confidence for normal-like synthetic samples and (ii) constraining overly deviated synthetic anomalies via configurable dual-space mixup, jointly mitigating anomaly shift while preserving realism.

\mypara{Mixup.}
Mixup interpolates samples and labels to regularize models. MixUp~\cite{zhang2017mixup} and its CV variants (CutMix~\cite{yun2019cutmix}, SnapMix~\cite{huang2021snapmix}, SmoothMix~\cite{jeong2021smoothmix}) enhance robustness by creating diverse yet label-faithful mixtures; manifold mixup~\cite{verma2019manifold} extends this to latent spaces, motivating our dual-space design. For time series, \cite{demirel2023finding} mixes phase and amplitude, and TimeMixer++~\cite{wang2024timemixer++} mixes across scales, but these methods mainly target classification robustness rather than KPI anomaly detection with injected anomalies.

\begin{table}[t]
\caption{Comparison of related KAD methods.}
\centering
\scalebox{0.76}{
\begin{threeparttable}
\begin{tabular}{l|cc|cc| c}
\hline
\bfseries Method & \bfseries N & \bfseries A$^{1}$ & \multicolumn{2}{c|}{\bfseries Anomaly Shift$^{2}$ } & \bfseries \shortstack{Covered structured\\TS anomaly type(s)$^{3}$}\\
\cline{4-5}
& & & Close & Far & \\
\hline
KAD-Disformer~\cite{yu2024pre} & $\checkmark$ & $\times$ & - & - & $\times$\\
ART~\cite{sun2024art} & $\checkmark$ & $\times$ & - & - & $\times$\\
RoCA~\cite{mou2025roca} & $\checkmark$ & $\times$ & - & - & $\times$ \\
NCAD~\cite{carmona2021neural} & $\checkmark$ & $\checkmark$ & $\times$ & $\times$ & G, Ctx\\
AnomalyBert~\cite{jeong2023anomalybert} & $\checkmark$ & $\checkmark$ & $\times$ & $\times$ & G, Ctx\\
GenIAS~\cite{darban2025genias} & $\checkmark$ & $\checkmark$ & $\checkmark$ & $\times$ & G, Ctx, Sh, Sea, Tr\\
RedLamp~\cite{obata2025robust} & $\checkmark$ & $\checkmark$ & $\checkmark$ & $\times$ & G, Ctx, Sh, Sea, Tr\\
\ourmethod & $\checkmark$ & $\checkmark$ & $\checkmark$ & $\checkmark$ & G, Ctx, Sh, Sea, Tr\\
\hline
\end{tabular}
\begin{tablenotes}
    \item[1] N and A indicate whether the method uses normal and abnormal samples.
    \item[2] ``Close'' means addressing normal-like synthetic anomalies, and ``Far'' means avoiding generating excessively deviated outliers.
    \item[3] Abbrev.: G=Global, Ctx=Contextual, Sh=Shapelet, Sea=Seasonal, Tr=Trend.
\end{tablenotes}
\end{threeparttable}
}
\label{relate_work}
\end{table}


\section{Our Approach}


\subsection{Problem Definition}
We formulate the monitored KPIs as time series.
Given an ordered time series $\mathcal{S}=\left\{x_1, x_2,\dots,x_l\right\}$ collected during an $l$-length time period.  $x_{i}\in \mathop{\mathbb{R}}^{d}$ is a $d$-dimensional vector collected at timestamp $i$. 
The time series will be denoted as univariate if $d=1$, and as multivariate if $d>1$. 
Conventional approaches typically split the long time series $\mathcal{S}$ into a set of time subsequences, i.e., $\mathcal{D}=\left\{{\bf X}_1,{\bf X}_2,\dots,{\bf X}_N\right\}$ by sliding windows whose length is set to be $t$. 
A sample ${\bf X}_i=\left\{x_1,x_2,\dots,x_t\right\}$ is the collection of points within a time subsequence, and $N$ is the number of samples. 
Time step $\delta \leq t$ is the stride of sliding, where the samples are overlapping if $\delta<t$.
To better describe the characteristics of pattern-wise anomalies, we adopt structural modeling \cite{lai2021revisiting} to represent a time-series sample as $X =  \Gamma(2\pi\omega T) + \Theta(T)$, where $T=\left\{1, 2,\dots,t\right\}$.
$\Gamma$ deﬁnes the basic shapelet function, which not only describes the shape of the time series but also includes relationships between various dimensions.
These relationships can be depicted by a covariance matrix $cov(X, X)$.
$\omega$ is the seasonality and $\Theta$ is the trend function describing the direction of $X_i$.
Correspondingly, $\mathcal{D}$ has a set of labels $\mathcal{Y}=\left\{y_1, y_2,\dots, y_N\right\}$ with $y_i\in \{0,1\}$ indicating that the sample ${\bf X}_i$ is normal (0) or anomalous (1).
The goal is to predict a label $\hat{y}_i\in \{0,1\}$ given a time series ${\bf X}_i$. 
We calculate an anomaly score ${\bf S}_{i}$  instead of giving the binary labels directly, and a predicted label can be obtained by comparing ${\bf S}_{i}$ to a predefined threshold $\tau$.

\begin{figure*}[t]
\centering
\includegraphics[width=0.98\linewidth]{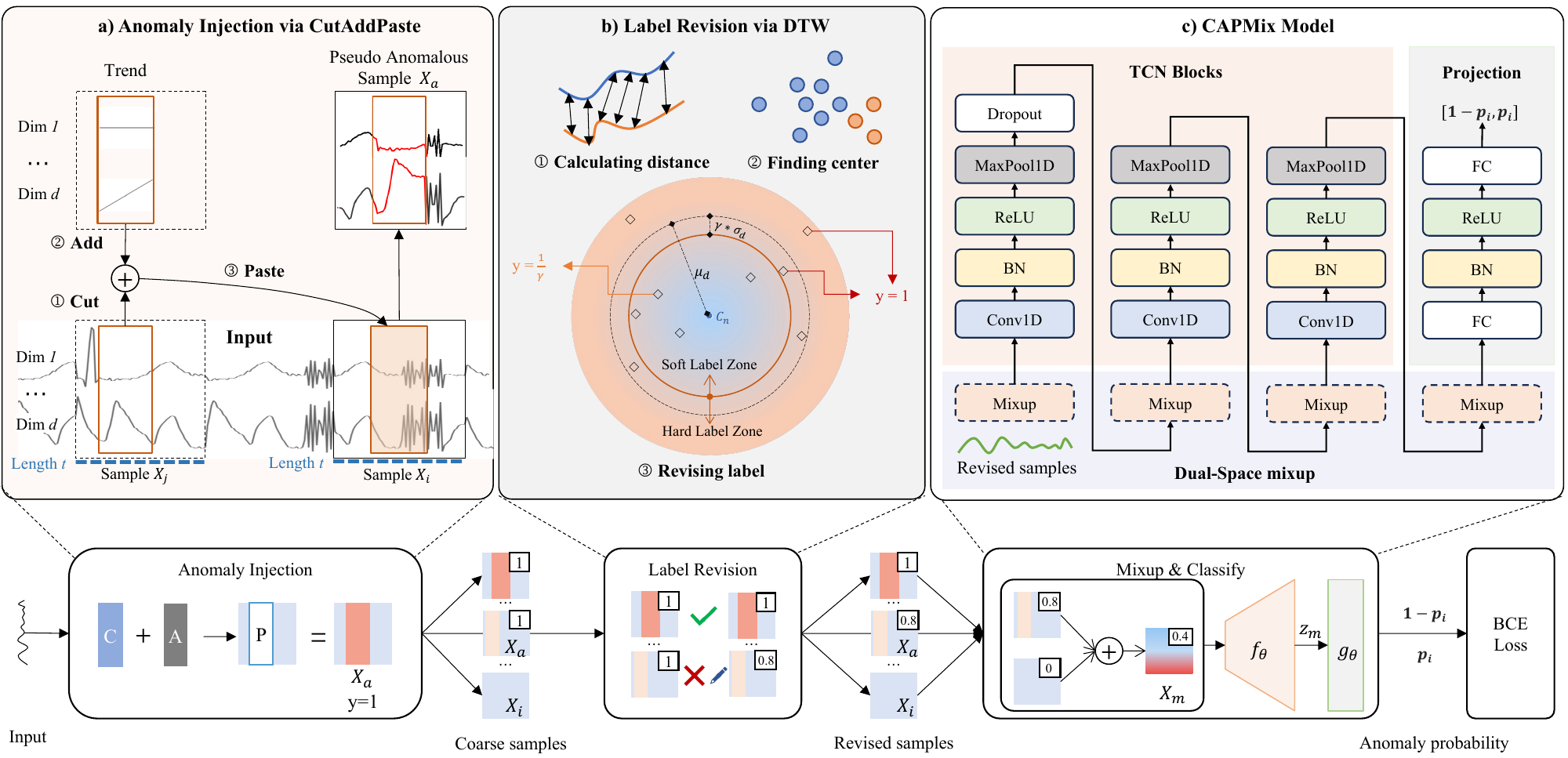}
\vspace{-0.4em}
\caption{Overview of \ourmethod. (a) CutAddPaste generates diverse synthetic anomalies by injecting patches into normal sequences. (b) Label Revision adjusts pseudo labels using DTW distance to the normal center, mitigating normal-like generations. (c) Dual-Space Mixup performs mixup in both input and latent spaces to learn robust, consistent decision boundaries.}
\label{fig:overview}
\end{figure*}

\subsection{Overview}
To effectively leverage anomaly knowledge and reduce the influence of benign fluctuations in noisy and dynamic environments, we propose \ourmethod for KPI anomaly detection. After injecting structured anomalies into time-series windows, \ourmethod mitigates the anomaly-shift issue in anomaly-assumption training through two key insights: (i) label revision prevents the few normal-like synthetic samples from being assigned hard anomaly labels; and (ii) dual-space mixup blends synthetic anomalies with normal samples in both input and latent spaces, preventing unrealistic anomalies from distorting the decision boundary.
As shown in Fig.~\ref{fig:overview}, \ourmethod maps multivariate time-series windows to anomaly scores via a left-to-right pipeline: anomaly injection, label revision, dual-space mixup with TCN-based representation learning, and classification.
Starting from raw time-series windows, we inject sparse but informative synthetic anomalies using the controlled base method.
These samples are then processed by the Label Revision (LR) module based on Dynamic Time Warping (DTW) distance to reduce the impact of anomaly shift.
Finally, the revised samples are fed into Dual-Space (DS) mixup and TCN blocks to learn robust representations in both input and latent spaces, and the classifier outputs anomaly probabilities optimized with Binary Cross-Entropy (BCE) loss.

\subsection{\ourmethod Pipeline}

\subsubsection{Base Anomaly Injection}

We adopt CutAddPaste~\cite{wang2024cutaddpaste}---a generic time-series anomaly augmentation method---as the base injection module in \ourmethod. It extends CutPaste~\cite{li2021cutpaste} to time series and generates anomalies by cutting a patch from a source window, optionally adding a trend term, and pasting it into a destination window. While effective, naive cut-add-paste may introduce abrupt boundaries, potentially aggravating anomaly shift.

\begin{figure}[htbp]
\centering
\includegraphics[width=\linewidth]{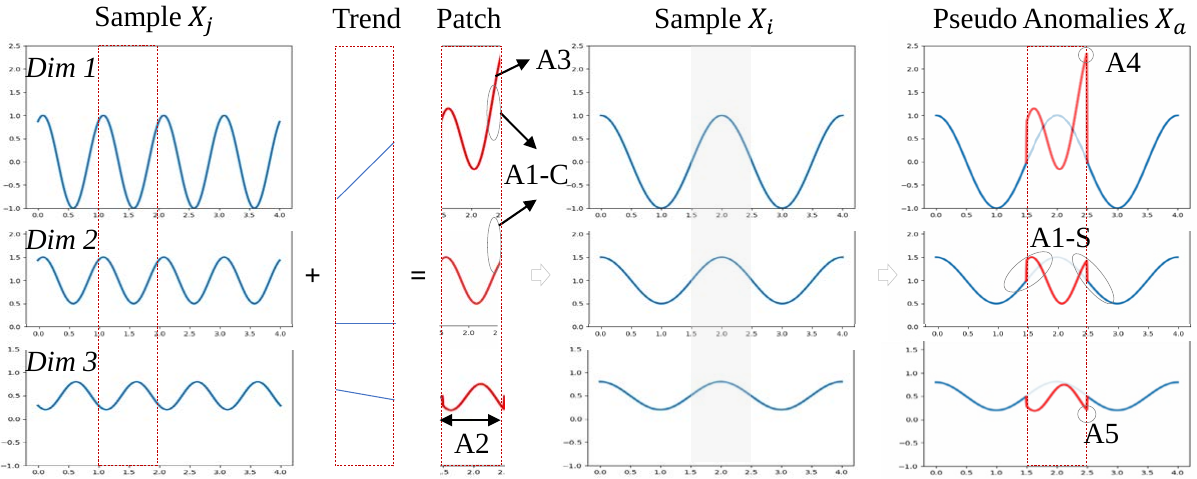} 
\caption{Example of generating an anomalous sample over sine waves. The original $Dim 2 = 0.5\cdot Dim 1+1$. There are pattern-wise anomalies in terms of shapelet (A1-S: shape, A1-C: correlation), seasonality (A2), and trend (A3). Point-wise anomalies include global (A4) and contextual (A5) ones. }
\label{fig:cutaddpaste_process}
\end{figure} 

\begin{itemize}[leftmargin=1.2em]
\item \textbf{Cut:} sample a patch of length $r\ge\zeta$ from another arbitrary window $X_j$.
\item \textbf{Add:} add a linear trend term $V_{trend}$ optionally on selected dimensions.
\item \textbf{Paste:} replace values at a random position in $X_i$ with the patched segment.
\end{itemize}

Formally, given the paste (destination) sample ${\bf X}_i= \Gamma_{i}(2\pi \omega_{i} T) + \Theta_{i}(T)$ and the cut (source) sample ${\bf X}_j= \Gamma_{j}(2\pi \omega_{j} T) + \Theta_{j}(T)$, where $T=\left\{1, 2,\dots,t\right\}$.
${\bf X}_i$ and ${\bf X}_j$ come from $\mathcal{S}$, which has been standardized.
The trend item $V_{trend}=\rho\cdot m \cdot R$, where $R=\left\{1, 2,\dots,r\right\}$.
$m=(m_{1},m_{2},\dots,m_{d})$ is a d-dimensional random slope vector, where $-1< m_{i}< 1$.
Hyperparameter $\rho>0$ controls the degree of the trend term.
After the above augmentation, the pseudo-abnormal sample ${\bf X}_a$ is defined as:$\left\{x_{b} | b \in T \right\}, $
and the formulation of $X_a$ involving ${\bf X}_i$, ${\bf X}_j$, and ${\bf V}_{trend}$ is depicted as follows:
\begin{equation}  
x_{b}=\left\{
\begin{array}{ll}
\Gamma_{i}(2\pi \omega_{i} b) + \Theta_{i}(b),     & b \in \left\{1,\dots,k_{i}-1\right\} \\
\Gamma_{j}(2\pi \omega_{j}(b+k)) + \Theta_{j}(b+k) & \multirow{2}{*}{$b \in \left\{k_{i},\dots,k^{'}+r \right\}$} \\
\hfill +\rho m (b-k^{'}),                     & \\
\Gamma_{i}(2\pi \omega_{i} b) + \Theta_{i}(b),     &  b \in \left\{k_{i}+r,\dots,t\right\} ,
\end{array} \right.
\label{Eq:pseudo_abnormal_point}
\end{equation}
where $k=k_{j}-k_{i}$, $k^{'}=k_{i}-1$. $k_i<t-r$ and $k_j<t-r$ are the random pasting position and cutting position, respectively.
As illustrated in Fig.~\ref{fig:cutaddpaste_process}, considering the subsequence $X[t_{1}:t_{2}]$, i.e., $\left\{x_{t_{1}}, x_{t_{1}+1},\dots,x_{t_{2}}\right\}$, we formulate the following relevant five types of anomalies onto $X_a$:
\begin{itemize}[leftmargin=1.2em]
\item \textbf{A1-S: Shape.}
Generally, $\Gamma_i$ is different from $\Gamma_j$ because $X_j$ is chosen randomly.
Therefore, an arbitrary sub-sequence $X_{a}[t_{1}:t_{2}]$ from $t_{1} \in [1:k_{i})$ to $t_{2} \in [k_{i}+1, T]$ contains at least one hop transition of the shape function $\Gamma$, forming the shape anomalies.

\item \textbf{A1-C: Correlation.} For $X_{i}$, the original relationships among the multiple variables are defined as $cov(X_i,X_i)={\rm E}[(X_i-{\rm E}[X_{i}])([(X_i-{\rm E}[X_{i}])^T]$.
Because the $i$-th slope $m_i$ is an independently generated random number, ${\rm E}[X_{a}]$ will deviate from ${\rm E}[X_{i}]$, further causing $cov(X_a,X_a)\neq cov(X_i,X_i)$, i.e., correlation anomalies.

\item \textbf{A2: Seasonality.} The seasonality parameter of the patch $X_a[k_{i}:k_{i}+r-1]$ is $\omega_j$ rather than the expected $\omega_i$.

\item \textbf{A3: Trend.} After the augmentation, the trend function of the patch $\Theta_a = \Theta_j + V_{trend}$, which is different from the expected $\Theta_i$.

\item \textbf{A4 \& A5: Point-wise.} 
When $b=k^{'}+r$, $x_{b}=X_{j}[k_{j}+r-1]+\rho mr$. If $m_{i} \neq 0$, such as $Dim 1$ and $Dim 3$ in Fig.~\ref{fig:cutaddpaste_process}, $x_{b}$ is proportional to $r$.  Therefore, when $r$ is large enough, $|x_{b}-\mu|$ will exceed the $3\sigma$ to create a global (A4) or contextual (A5) anomaly, where $\mu$ and $\sigma$ are the mean and standard deviation of the whole time series $\mathcal{S}$ or the neighborhood points.
\end{itemize}

\subsubsection{Label Revision via DTW}

KPI time series naturally exhibit benign fluctuations such as periodic oscillations, workload-driven jitter, and transient spikes. When we inject synthetic anomalies, such ``normal-like'' oscillations can cause an anomaly-shift effect: some injected samples may not present salient abnormal signatures, yet are still assigned one-hot anomaly labels. Such coarse supervision may blur the decision boundary and encourage over-sensitive detectors that misinterpret normal KPI turbulence as incidents.

To mitigate this label uncertainty, we propose a DTW-based label revision mechanism. For each generated sample, we compute its DTW distance to a normality center and assign a soft anomaly confidence in $[0,1]$ rather than a hard one-hot label, so that samples close to normal KPI dynamics receive weaker anomaly supervision.

We first estimate the normality center $C_n$ of the training data, assuming that the majority is normal. Given a collection of normal sequences $\mathcal{D}=\left\{{\bf X}_1,{\bf X}_2,\dots,{\bf X}_N\right\}$, the center is computed as the mean sequence over all time steps:
\begin{equation}
    C_n = \frac{1}{N}\sum_{i=1}^{N}X_i.
\label{Eq:center_calculation}
\end{equation}
For each generated sample $X_a$, we compute its DTW distance $d(X_a,C_n)$ to $C_n$. Denote the mean and standard deviation of distances from the original training samples to $C_n$ as $\mu_d$ and $\sigma_d$. We then define a soft-label region controlled by the threshold $\mu_d + \gamma\sigma_d$:
\begin{equation}
    d(X_a,C_n) \le \mu_d + \gamma\sigma_d,
\end{equation}
where $\gamma\ge 0$ adjusts the tolerance to normal KPI oscillations. Samples within this region are treated as ambiguous and assigned a soft anomaly label $y_r = 1/\gamma$. Samples outside the region are considered confidently abnormal and assigned a hard label $y_r=1$. Then the original label set is refined as $\mathcal{Y}_r$. Formally, each label $y_r$ in $\mathcal{Y}_r$ is:
\begin{equation}
y_{r}= \begin{cases}1, & \text{  if  }    d\left(X_{\mathrm{syn}, C_{n}}\right)>\mu_{d}+\gamma \sigma_{d} \\ \frac{1}{\gamma}, & \text{  otherwise.}\end{cases}
\label{Eq:label_revision}
\end{equation}
Although one may assign continuous soft labels using a distance ratio (e.g., $d/(\mu_d+\gamma\sigma_d)$), this introduces extra computation and may yield unstable supervision due to scale sensitivity. We therefore use the controlled label $1/\gamma$ to balance robustness and training clarity.

\subsubsection{Dual-Space Mixup Integrated TCN}

Beyond mitigating under-distinctive synthetic anomalies via label revision, we also observe the opposite failure mode in AIOps KPI augmentation: injected anomalies may become \emph{over-distinctive} and thus unrealistic. For instance, they may exhibit abrupt level shifts or extreme spikes that rarely occur in production KPIs. Training on such samples can bias the detector toward exaggerated signatures and reduce its sensitivity to subtle but operationally critical KPI degradations. To address this, we introduce a Dual-Space Mixup strategy within a TCN backbone, which regularizes the learned boundary in both the input space and the latent feature space.

The proposed \ourmethod model employs a three-block TCN to convert a $t$-length sequence $X_i$ into a fixed-size representation $z_i$. It is implemented by three temporal convolutional blocks $f^{(l)}(\cdot)$, where $l \in [1,2,3]$. Each block contains a Conv1D layer, a Batch Normalization (BN) layer, a ReLU activation function, and a MaxPool1D layer; the first block additionally includes Dropout. We perform mixup at configurable depths. For layer $l$, we mix intermediate features and their corresponding labels as:
\begin{equation}
\begin{aligned}
X^{(l)}_{Mix} &= \lambda \cdot f^{(l)}(X_i) + (1-\lambda) \cdot f^{(l)}(X_j), \\
y^{(l)}_{Mix} &= \lambda \cdot y^{(l)}_i + (1-\lambda) \cdot y^{(l)}_j,
\end{aligned}
\label{Eq:mixup_layer_l}
\end{equation}
where $y^{(l)}$ denotes the label associated with the $l$-th mixup block. In particular, setting $l=0$ in Eq.~(\ref{Eq:mixup_layer_l}) yields the standard input-space mixup, i.e., directly mixing the raw inputs $X$ and labels $y$.
This layer-wise design enables switching mixup on/off at different levels, making the framework extensible to different KPI characteristics and anomaly types.

The representation ${z_i}$ is further fed into a learnable nonlinear projector $g_{\theta}:\mathcal{C}\mapsto\mathcal{Q}$ to obtain the projection $q_i$. The projector is an MLP with one hidden layer using BN and ReLU, mapping encoder features into a 2-dimensional projection space. The training objective is the binary cross-entropy loss, detailed in Section~\ref{sec:Objective}.

Let $\hat {\mathcal{D}_{syn}}$ denote the revised synthetic anomaly distribution after applying DTW-based soft labeling and DS-Mixup. Intuitively, these two components jointly reduce the distribution gap between synthetic and real anomalies, i.e.,
\begin{equation}
    Dist(\mathcal{D}_{real}, \hat {\mathcal{D}_{syn}})  < Dist(\mathcal{D}_{real}, {\mathcal{D}_{syn}}),
\end{equation}
which we further validate via visualization in our experiments.

\subsubsection{Putting LR and DS-Mixup Together}

Label Revision (LR) via DTW addresses the \emph{low-distinctiveness} side of anomaly shift: due to benign KPI oscillations (e.g., periodicity and workload-driven jitter), some injected samples may remain close to normal dynamics, and hard anomaly labels can be overconfident. LR uses DTW distance to the normality center to calibrate such samples with soft labels, reducing misleading supervision.
DS-Mixup complements LR by addressing the \emph{over-distinctiveness} side: it constrains synthetic anomalies that deviate excessively from plausible KPI behaviors. Input-space mixup generates hybrid KPI segments that preserve realistic local dynamics, while latent-space mixup interpolates intermediate representations and soft labels to encourage smooth transitions in feature space. Together, these operations promote smoother decision boundaries and improve generalization across diverse KPI fluctuations and anomaly patterns.

We will validate the contribution of LR and DS-Mixup through ablation experiments in Section~\ref{sec:ablation}.

\begin{algorithm}[t]
\caption{\ourmethod Pipeline}
\label{pseudo}
\begin{algorithmic}[1]
\Require Subsequence set $\mathcal{D} = \{{\bf X}_i\}_{i=1}^N$, initial labels $\mathcal{Y}=\{y_i\}_{i=1}^N$. Parameters: $\rho, \zeta, e, \gamma, v$.
\Ensure Optimized networks $f, g$.

\State \textbf{Stage 1: Adaptive Data Augmentation}
\For{each ${\bf X}_i \in \mathcal{D}$}
    \State $r \gets \max(\zeta, \text{randint}(1, t))$, $pos_{c}, pos_{p} \gets \text{randint}(1, t-r)$;
    \State Extract ${\bf X}_{sub} \gets {\bf X}_{j}[pos_{c} : pos_{c}+r]$ from random $j \in [1, N]$;
    \State Select $e$ dimensions $\mathcal{D}_{sub} \subset \{1, \dots, d\}$;
    \State ${\bf X}_{sub}[:, \mathcal{D}_{sub}] \gets {\bf X}_{sub}[:, \mathcal{D}_{sub}] + \text{Trend}(\rho)$ \Comment{Inject drift on $e$ dimensions};
    \State Generate augmented ${\bf X}'_i$ by pasting ${\bf X}_{sub}$ into ${\bf X}_i$ at $pos_p$;
\EndFor

\State \textbf{Stage 2: Label Revision \& Refinement}
\State $\mathcal{D}' \gets \text{RandomSample}(\{{\bf X}'_i\}, v \cdot N)$, set initial labels $\mathcal{Y}' = \{1\}^N$;
\State Compute normal center $C_n = \text{Eq.}(\ref{Eq:center_calculation})$ using $\mathcal{D}$;
\State Revise labels $y'_i \in \{1, 1/\gamma\}$ for ${\bf X}'_i \in \mathcal{D}'$ via Eq. (\ref{Eq:label_revision});
\State $\mathcal{D}_{total} \gets \mathcal{D} \cup \mathcal{D}'$, $\mathcal{Y}_{total} \gets \mathcal{Y} \cup \mathcal{Y}'$;

\State \textbf{Stage 3: Joint Optimization with DS-Mixup}
\While{not converged}
    \State Sample a mini-batch $\{{\bf X}_b, y_b\} \subset \{\mathcal{D}_{total}, \mathcal{Y}_{total}\}$;
    \State Forward pass with DS-Mixup via Eq. (\ref{Eq:mixup_layer_l});
    \State Compute $\mathcal{L}$ via Eq. (\ref{Eq:cross_entropy}) and update $f, g$;
\EndWhile
\Return $f, g$
\end{algorithmic}
\end{algorithm}

\subsection{Model Training Objective}
\label{sec:Objective}

The loss is obtained by calculating the projection $q_i$ and the label ${y}_i$.
Inspired by hypersphere classiﬁcation (HSC) \cite{hendrycks2018deep,ruff2020rethinking}, we use the binary cross-entropy loss as the training objective. 

The projector $g_{\theta}$ outputs the 2-dimensional projections $\mathcal{Q}=\left\{q_1,...,q_i,...,q_N\right\}$, and $softmax$  maps $q_i$ to a pair of probabilities $[1-p_i, p_i]$, where $p_i$ and $1-p_i$ represent the probability of being anomalous and being normal, respectively. We use $softmax$ instead of $sigmoid$ to obtain the two probabilities, separately.
The binary cross-entropy loss is defined as:
\begin{equation}
\mathcal{L} = 
-\frac{1}{N}\sum_{i=1}^{N}[y_i\cdot log(p_i) + (1-y_i)\cdot log(1-p_i)],
\label{Eq:cross_entropy}
\end{equation}
where $y_i$ is the label in $\mathcal{Y}_r$. 
For the training set with labels, $\mathcal{Y}_r$ contains the original labels annotated by domain experts and our labels corresponding to the pseudo-anomalous samples. 
It is worth mentioning that our method does not require the training set to contain anomalies.
As shown in Table~\ref{results}, \textsc{\ourmethod} performs well on UCR, SWaT, and WADI, all of which are generic training sets without anomalies. The pseudo-code of \ourmethod in Pytorch style is provided in Algorithm \ref{pseudo}. 

In the test phase, we consider the probability $p_i$ of the subsequence sample ${\bf X}_i$ as the anomaly score $S_{i}\in[0,1]$.
Then, we use the following standard to determine whether ${\bf X}_i$ can be classified as anomalous:
\begin{equation}  
x_{i}=\left\{
\begin{array}{lcl}
anomaly,& & S_{i} > \tau\\
 normal,& & S_{i} \le \tau \quad ,
\end{array} \right.
\label{ADdetection}
\end{equation}
where $\tau$ is a predefined threshold.

\section{Experiments}
\label{sec:experiments}
We systematically evaluate \ourmethod on multiple real-world KPI anomaly detection datasets to answer the following research questions. 

\begin{itemize}[leftmargin=1.2em]
    \item RQ1: What is the overall performance of \ourmethod?
    \item RQ2: How do the key components of \ourmethod contribute to performance and alleviate anomaly shift?
    \item RQ3: How robust is \ourmethod under varying contamination/noise ratios?
    \item RQ4: How do the key hyperparameters of \ourmethod affect its performance?
\end{itemize}
The code is available at \texttt{\url{https://github.com/alsike22/CAPMix}}.

\begin{table}[t]
\caption{Datasets description.}
\label{tab:datasets}
\centering
\small
\renewcommand{\arraystretch}{1.05}
\begin{tabular}{lcclrc}
\toprule
Dataset & \#Ent & \#Var & Domain & Points & Tra-Ano \\
\midrule
AIOps~\cite{KPI} & 29 & 1 & Cloud KPIs & 203{,}316 & 3.86\% \\
ASD~\cite{li2021multivariate} & 12 & 19 & Application & 154,171 & 0\% \\
SMD~\cite{su2019robust, li2021multivariate} & 28 & 38 & Server & 608,342 & 0\% \\
Exathlon~\cite{jacob2020exathlon} & 8 & 19 & Spark & 141,074 & 0\% \\
\midrule
UCR~\cite{wu2021current} & 250 & 1 & Various & 1{,}383{,}502 & 0\% \\
SWaT~\cite{mathur2016swat} & 1 & 51 & Waterworks & 63{,}441 & 0\% \\
WADI~\cite{ahmed2017wadi} & 1 & 127 & Waterworks & 61{,}993 & 0\% \\
ESA~\cite{kotowski2024european} & 1 & 6 & Satellite & 24{,}547{,}200 & 0.6\% \\
\bottomrule
\end{tabular}
\end{table}

\subsection{Experiment Setup}
\subsubsection{Datasets} 

The anomaly detection evaluation is conducted over KPI metrics datasets including  AIOps~\cite{KPI}, ASD~\cite{li2021multivariate}, SMD~\cite{su2019robust, li2021multivariate}, and Exathlon~\cite{jacob2020exathlon}.
On the other hand, time-series datasets from other domains are also included to verify the generalizability and scalability of \ourmethod: UCR~\cite{wu2021current}, SWaT~\cite{mathur2016swat}, WADI~\cite{ahmed2017wadi}, and ESA~\cite{kotowski2024european}  datasets.

Table~\ref{tab:datasets} summarizes the datasets and our preprocessing/splitting protocol. We segment each time series into sliding windows of length $t$ with step $\delta$. Here, Ent denotes the number of subsets/entities, Var the number of variables, Points the total number of data points, and Tra-Ano the proportion of anomalies in the training set. AIOps shows non-trivial training contamination (Tra-Ano $>0$), aligning with the unsupervised setting.

\begin{table*}[ht]
\renewcommand{\arraystretch}{1.15}
\centering
\small
\caption{Main results$^1$.}
\setlength{\tabcolsep}{1.5mm}
\scalebox{1.05}{%
\begin{threeparttable}
\begin{tabular}{c|cccc|cccc|c}
\toprule
\multirow{2}{*}{\textbf{Method}} & \multicolumn{4}{c|}{\textbf{KPI benchmarks}} & \multicolumn{4}{c|}{\textbf{Other time-series domain}} & \multirow{2}{*}{\textbf{Average}}\\
\cmidrule(lr){2-5} \cmidrule(lr){6-9}
& AIOps & SMD & ASD & Exathlon & UCR & SWaT & WADI & ESA & \\
\midrule
OC-SVM~\cite{scholkopf1999support} & 23.65 & 11.37 & 24.05 & 31.52 & 16.42 & 0.02 & 0.03 & 29.79 & 17.11\\
IF~\cite{liu2012isolation} & 3.86$\pm$0.07 & 6.07$\pm$0.32 & 16.82$\pm$0.72 & 44.05$\pm$1.65 & 4.51$\pm$0.44 & 12.37$\pm$2.39 & 0.88$\pm$0.13 & 0.09$\pm$0.00 & 11.08 \\
RRCF~\cite{guha2016robust} & 2.89$\pm$0.06 & 4.24$\pm$0.15 & 10.74$\pm$0.67 & 15.46$\pm$1.16 & 5.91$\pm$0.67 & 0.99$\pm$0.05 & 0.95$\pm$0.09 & 6.43$\pm$1.72 & 5.95\\
SR$^2$~\cite{ren2019time} & 8.52 & -- & -- & -- & 22.00 & - & - & - & 15.26\\
DAMP~\cite{lu2022matrix} & 0.18 & 4.29 & 4.87 & 16.87 & 32.26 & 0.70 & 0.18 & 0.09 & 7.43\\
\midrule
LSTM-ED~\cite{malhotra2016lstm} & 14.05$\pm$0.52 & 31.08$\pm$0.43 & 24.09$\pm$1.11 & 51.64$\pm$0.28 & 18.12$\pm$0.62 & 4.44$\pm$2.22 & 3.86$\pm$0.52 & 54.80$\pm$2.92 & 25.26\\
Deep SVDD~\cite{ruff2018deep} & 22.36$\pm$8.05 & 46.93$\pm$4.02 & 33.83$\pm$5.16 & 56.05$\pm$1.83 & 47.54$\pm$2.33 & 13.57$\pm$10.35 & 6.59$\pm$2.10 & 4.00$\pm$0.72 & 28.86\\
AOC~\cite{mou2023deep} & 39.82$\pm$4.95 & \textbf{60.85$\pm$0.29} & 35.11$\pm$1.07 & 61.59$\pm$0.44 & 21.39$\pm$0.80 & 27.59$\pm$0.00 & 0.36$\pm$0.00 & \underline{55.90$\pm$2.64} & 37.83\\
TCC~\cite{sohn2020learning,eldele2021time} & 3.28$\pm$1.36 & 5.54$\pm$1.22 & 9.83$\pm$0.83 & 59.54$\pm$19.35 & 0.50$\pm$0.09 & 16.71$\pm$8.80 & 2.09$\pm$4.03 & 25.30$\pm$21.00 & 15.35\\
AnoTrans~\cite{xu2021anomaly} & 0.58$\pm$0.51 & 17.49$\pm$6.79 & 5.65$\pm$6.01 & 23.03$\pm$12.16 & 6.38$\pm$2.56 & 33.07$\pm$1.75 & 1.68$\pm$0.02 & 1.13$\pm$0.20 & 11.13\\
MixMamba~\cite{alkilane2024mixmamba} & 16.72$\pm$0.13 & 27.72$\pm$0.27 & 31.54$\pm$0.81 & 44.30$\pm$0.53 & 12.88$\pm$0.54 & 2.42$\pm$0.07 & 3.68$\pm$0.68 & 25.20$\pm$5.47 & 20.56\\
MTSCAD~\cite{si2023beyond} & 44.85$\pm$1.24 & 51.70$\pm$0.70 & 27.22$\pm$0.73 & \underline{63.34$\pm$0.37} & 10.04$\pm$0.92 & 19.70$\pm$0.40 & 5.01$\pm$2.50 & 51.64$\pm$5.82 & 34.19\\
RoCA~\cite{mou2025roca} & 50.36$\pm$5.12 & 10.04$\pm$4.72 & 18.88$\pm$10.23 & 45.26$\pm$9.71 & 55.49$\pm$2.47 & 30.31$\pm$3.63 & 14.68$\pm$6.68 & 31.41$\pm$22.64 & 32.05\\
SHUE~\cite{feng2024sensitivehue} & 11.01$\pm$2.47 &	22.89$\pm$3.48	& 12.25$\pm$1.51	& 52.5$\pm$1.96	& 10.22$\pm$1.12	& 30.14$\pm$3.67	& \textbf{35.57$\pm$0.84}	& 16.36$\pm$4.93 & 23.87 \\
CATCH~\cite{wucatch} &	25.63$\pm$0.91	& 57.51$\pm$1.07 & 25.89$\pm$0.15	& 56.00$\pm$0.06	& 0.06$\pm$0.00	& 2.72$\pm$0.01	&0.03$\pm$0.00	& 16.46$\pm$0.00 & 23.04\\
MOC~\cite{wang2026moc}& 53.66$\pm$3.76	& 38.58$\pm$10.09	& 36.77$\pm$5.01	& 41.61$\pm$5.23	& 7.18$\pm$2.40 	& 27.19$\pm$2.14	& 14.14$\pm$8.46	&3.07$\pm$3.51 & 27.78 \\
\midrule
NCAD~\cite{carmona2021neural} & 41.06$\pm$3.32 & 16.49$\pm$3.07 & 32.14$\pm$7.64 & 24.55$\pm$2.94 & 22.24$\pm$2.99 & 7.54$\pm$2.48 & 6.84$\pm$2.53 & 22.03$\pm$1.07 & 21.61\\
AnomalyBert~\cite{jeong2023anomalybert} & 23.99$\pm$1.22 & 19.54$\pm$7.15 & 23.75$\pm$6.85 & 45.49$\pm$8.53 & 2.77$\pm$0.12 & 13.36$\pm$2.21 & 11.79$\pm$1.56 & 6.26$\pm$5.50 & 18.37\\
RedLamp~\cite{obata2025robust} &	23.21$\pm$1.26 & \underline{54.72$\pm$0.97}&\textbf{69.59$\pm$1.33}&	63.4$\pm$0.61	& 1.41$\pm$0.08 &	27.6$\pm$0.03	& 4.29$\pm$0.00 	&45.59$\pm$12.93 & 36.23 \\
CutAddPaste~\cite{wang2024cutaddpaste} & \underline{77.44$\pm$0.86} & 22.19$\pm$11.84 & 38.20$\pm$2.59 & 23.80$\pm$6.89 & \underline{69.98$\pm$1.44} & \underline{43.43$\pm$4.86} & 26.55$\pm$5.66 & 18.56$\pm$2.70 & \underline{40.02}\\
\ourmethod & \textbf{80.45$\pm$0.43} & 27.42$\pm$12.50 & \underline{61.30$\pm$1.96 }& 	\textbf{78.89$\pm$1.61} & \textbf{71.89$\pm$1.89} & \textbf{47.04$\pm$2.70} & \underline{34.08$\pm$6.67} & \textbf{84.46$\pm$5.50} & \textbf{60.69}\\
\bottomrule
\end{tabular}
\begin{tablenotes}
    \item[1] Average Best RPA F1-score (\%) with standard deviation for baselines and our method on datasets over 10 runs. The best results are in bold, and the suboptimal ones are underlined. 
    \item[2] SR is a univariate AIOps solution, yet it cannot be applied to multivariate time series anomaly detection.
\end{tablenotes}
\end{threeparttable}%

}
\label{results}
\end{table*}

\subsubsection{Metrics}

We evaluate anomaly detection using Revised Point-Adjusted (RPA) F1~\cite{hundman2018detecting}, which avoids the underestimation of point-wise (PW) metrics and the overly optimistic behavior of point-adjusted (PA) metrics~\cite{xu2018unsupervised}. Following common practice, we report the \emph{best} F1 score (Best RPA-F1) by sweeping the decision threshold over anomaly scores, which reflects a model/method's intrinsic ability to separate normal and anomalous patterns. In practical deployment, anomaly scores can also be converted to alarms via unsupervised thresholding methods (e.g., POT). For datasets with multiple subsets (AIOps and UCR), we report a weighted average over subsets, ${\rm F1_{entire}} = \sum_{i=1}^{M}\frac{e_{i}}{E}{\rm F1}_{i}$, where $M$ is the number of subsets, $E$ is the total number of anomalies, and $e_i$ is the anomaly count in subset $i$.

\subsubsection{Baselines}

We compare \ourmethod with representative baselines, covering traditional methods, deep detectors built on the normality assumption, and anomaly-assumption/injection approaches.

\textit{Traditional AD baselines.}
We include One-Class SVM (OC-SVM)~\cite{scholkopf1999support}, Isolation Forest (IF)~\cite{liu2012isolation}, Robust Random Cut Forest (RRCF)~\cite{guha2016robust}, Spectral Residual (SR)~\cite{ren2019time}, and DAMP~\cite{lu2022matrix}. For the large ESA dataset, these baselines are adapted to handle long windows. 

\textit{Normality-assumption baselines.}
We evaluate deep detectors under single assumptions---reconstruction, one-class classification, or prediction---including LSTM-ED~\cite{malhotra2016lstm}, Deep SVDD~\cite{ruff2018deep}, Anomaly Transformer~\cite{xu2021anomaly}, MixMamba~\cite{alkilane2024mixmamba}, MTSCAD~\cite{si2023beyond}, SensitiveHUE (SHUE)~\cite{feng2024sensitivehue}, CATCH~\cite{wucatch}, and MOC~\cite{wang2026moc}. 
We also include fused/multi-assumption baselines: RoCA~\cite{wang2023deep}, AOC~\cite{mou2023deep}, and TCC~\cite{sohn2020learning,eldele2021time}.

\textit{Anomaly-assumption baselines.}
We consider NCAD~\cite{carmona2021neural}, AnomalyBert~\cite{jeong2023anomalybert},RedLamp~\cite{obata2025robust}, and CutAddPaste~\cite{wang2024cutaddpaste}. 
For NCAD, we use its supervised setting on AIOps as it yields better performance. Following prior work~\cite{wang2023deep,mou2023deep,carmona2021neural}, Conv1D is adopted to migrate Deep SVDD-style objectives to time series. For TCC, we pre-train representations with TS-TCC~\cite{eldele2021time} and fine-tune using Deep SVDD principles.


\vspace{-0.2em}
\subsection{Main Results (RQ1)} 
We report the AD performance of the baselines and \ourmethod in Table~\ref{results}, in terms of F1 score. From top to bottom, the table is organized into three groups: traditional, normality assumption-based, and anomaly assumption-based methods, from which several key observations can be made.

First, on the KPI anomaly detection (KAD) benchmarks (AIOps, SMD, ASD, and Exathlon), \ourmethod consistently achieves strong RPA F1 scores and remains competitive across datasets with diverse KPI dynamics. In contrast, traditional and normality-assumption-based methods (e.g., OC-SVM, Deep SVDD, and AOC) show larger performance fluctuations, suggesting sensitivity to dataset-specific noise patterns and anomaly characteristics.
Second, on the remaining four datasets, the results indicate that \ourmethod also serves as a general-purpose time-series anomaly detector. In particular, multivariate settings remain challenging for most baselines, while \ourmethod provides consistent gains and achieves the best overall performance on SWaT/WADI and a strong result on ESA.

Across baselines, classic methods such as DAMP and SR, can be competitive on some univariate cases, whereas fused-normality approaches including RoCA and AOC are generally stronger than single-assumption deep models, highlighting the value of combining complementary normality signals. Finally, among anomaly-assumption-based methods, CutAddPaste and RedLamp are strong baselines, and \ourmethod further mitigates anomaly shift, yielding the best average performance across all eight datasets.

\begin{table*}[!ht]
\renewcommand{\arraystretch}{1.25}
\centering
\small
\caption{ Ablation results of average Best F1-score (\%) ) with standard deviation over 10 runs. 
} 

\centering
\setlength{\tabcolsep}{1.5mm}
\scalebox{1}{
\begin{threeparttable}
\begin{tabular}{c|ccc|cccc|cccc|c}
\toprule
\multirow{2}{*}{\textbf{Variant}} & \multicolumn{3}{c|}{\textbf{Components$^{1}$}} & \multicolumn{4}{c|}{\textbf{KPI benchmarks}} & \multicolumn{4}{c|}{\textbf{Other time-series domain}} & \multirow{2}{*}{\textbf{Average}}\\
\cmidrule(lr){2-4} \cmidrule(lr){5-8} \cmidrule(lr){9-12}
& C & L & M & AIOps & SMD & ASD & Exathlon & UCR & SWaT & WADI & ESA & \\
\midrule
NoAug & $\times$ & $\times$ & $\times$ & 74.25$\pm$2.60 & 15.58$\pm$11.74 & 15.56$\pm$6.50 & 23.80$\pm$6.89 & 59.08$\pm$2.42 & 21.64$\pm$8.42 & 7.90$\pm$6.44 & 15.33$\pm$7.84 & 29.14 (\textcolor{black}{$\Large \downarrow$ 31.55})\\
CAP & $\checkmark$ & $\times$ & $\times$ & 77.44$\pm$0.86 & 22.19$\pm$11.84 & 38.20$\pm$2.59 & 73.73$\pm$4.72 & 69.98$\pm$1.44 & 43.43$\pm$4.86 & 26.55$\pm$5.67 & 18.56$\pm$2.70 & 46.26 (\textcolor{black}{$\Large \downarrow$ 14.43})\\
CAP-$lr$ & $\checkmark$ & $\checkmark$ & $\times$ & 80.13$\pm$0.53 & 24.06$\pm$10.74 & 42.29$\pm$3.40 & 73.73$\pm$4.72 & 69.98$\pm$1.44 & 43.43$\pm$4.86 & 26.55$\pm$5.67 & 18.56$\pm$2.70 & 47.34 (\textcolor{black}{$\Large \downarrow$ 13.35})\\
CAP-$mix$ & $\checkmark$ & $\times$ & $\checkmark$ & 78.15$\pm$0.86 & 23.77$\pm$11.62 & 45.55$\pm$3.66 & 78.18$\pm$2.90 & 68.65$\pm$2.45 & 47.04$\pm$2.70 & 34.08$\pm$6.67 & 84.46$\pm$6.67 & 57.49 (\textcolor{black}{$\Large \downarrow$ 3.20})\\
\rowcolor{black!10} Total & $\checkmark$ & $\checkmark$ & $\checkmark$ & 80.45$\pm$0.43 & 27.42$\pm$12.50 & 61.30$\pm$1.96 & 78.89$\pm$1.61 & 71.89$\pm$1.89 & 47.04$\pm$2.70 & 34.08$\pm$6.67 & 84.46$\pm$6.67 & 60.69\\
\bottomrule
\end{tabular}
\begin{tablenotes}
    \item[1] C refers to the process of anomaly injection via CutAddPaste. L refers to label revision. M refers to dual-space mixup.
\end{tablenotes}
\end{threeparttable}
}
\label{tab:ablation}

\end{table*}

\subsection{Ablation Study (RQ2)}
\label{sec:ablation}

\subsubsection{Component Ablation.} 

We study four variants, ranging from a vanilla TCN to the full \ourmethod. Table~\ref{tab:ablation} reports RPA scores, where the Average column measures the relative drop with respect to the full model, revealing the contribution of each component.
\begin{itemize}[leftmargin=1.2em]
\item \textbf{NoAug.} TCN trained on original data with BCE loss.
\item \textbf{CAP.} Adds CutAddPaste-based anomaly injection.
\item \textbf{CAP-$lr$.} Further introduces label revision by assigning soft labels ($1/\gamma$) to near-normal synthetic anomalies.
\item \textbf{CAP-$mix$.} Applies dual-space mixup on injected samples without label revision.
\end{itemize}

Removing augmentation (\textit{NoAug}) yields the weakest results on all datasets, suggesting that learning from limited normal data alone is insufficient for robust anomaly detection in noisy settings; moving to \textit{CAP} consistently improves performance, with larger gains on KPI benchmarks such as AIOps and ASD, indicating that prior-guided anomaly injection is more effective than relying on raw noisy observations because it encourages the model to attend to structured deviation patterns rather than incidental fluctuations. Adding label revision (\textit{CAP-$lr$}) further improves over CAP, especially on KPI datasets, implying that calibrated supervision under noise stabilizes representation learning and reduces the impact of contaminated or ambiguous samples. Dual-space mixup (\textit{CAP-$mix$}) provides substantial benefits across domains, with particularly clear gains on challenging datasets such as ESA; compared with CAP and CAP-$lr$, DS improves generalization, suggesting that smoothing decision boundaries in both input and representation spaces enhances robustness to distribution shifts and noise. Finally, the full model achieves the best results on nearly all datasets, confirming that the three components are complementary: AJ and LR mainly strengthen robustness under noisy supervision, while DS further improves cross-domain generalization, yielding a more stable and reliable anomaly detection model.

\begin{figure}[ht]
    \centering
    \subfigure[AIOps]{%
        \begin{minipage}[t]{0.45\linewidth}
            \centering
            \includegraphics[width=\linewidth,height=0.75\linewidth]{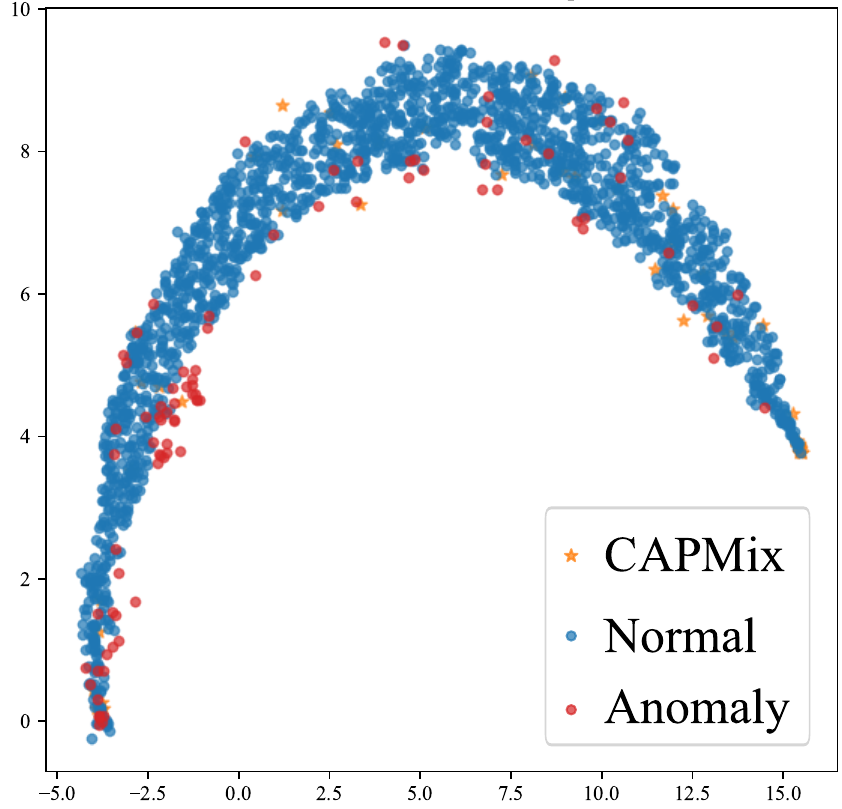}
            \label{fig:umap_capmix_kpi}
        \end{minipage}%
    }
    \subfigure[UCR]{%
        \begin{minipage}[t]{0.45\linewidth}
            \centering
            \includegraphics[width=\linewidth,height=0.75\linewidth]{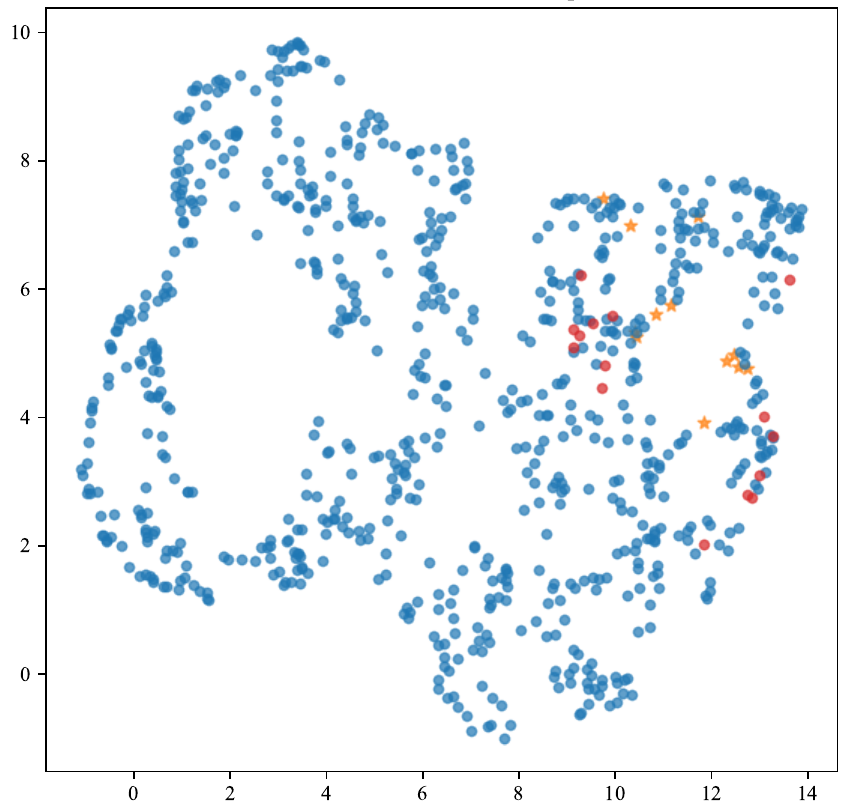}
            \label{fig:umap_capmix_ucr}
        \end{minipage}%
    }\\
    \subfigure[Exathlon]{%
        \begin{minipage}[t]{0.45\linewidth}
            \centering
            \includegraphics[width=\linewidth,height=0.75\linewidth]{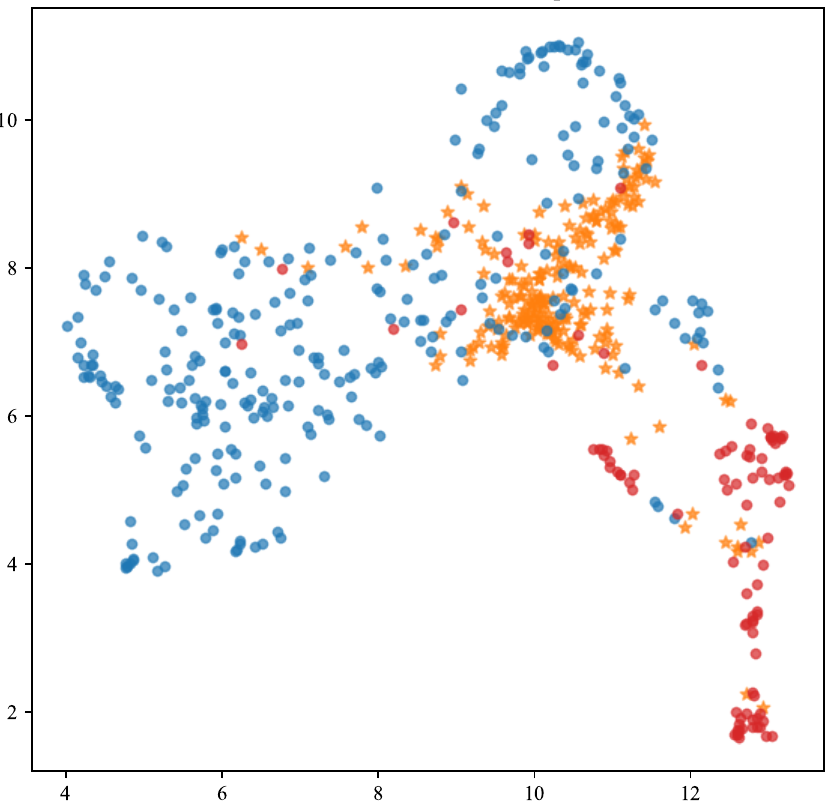}
            \label{fig:umap_capmix_wadi}
        \end{minipage}%
    }
    \subfigure[ESA]{%
        \begin{minipage}[t]{0.45\linewidth}
            \centering
            \includegraphics[width=\linewidth,height=0.75\linewidth]{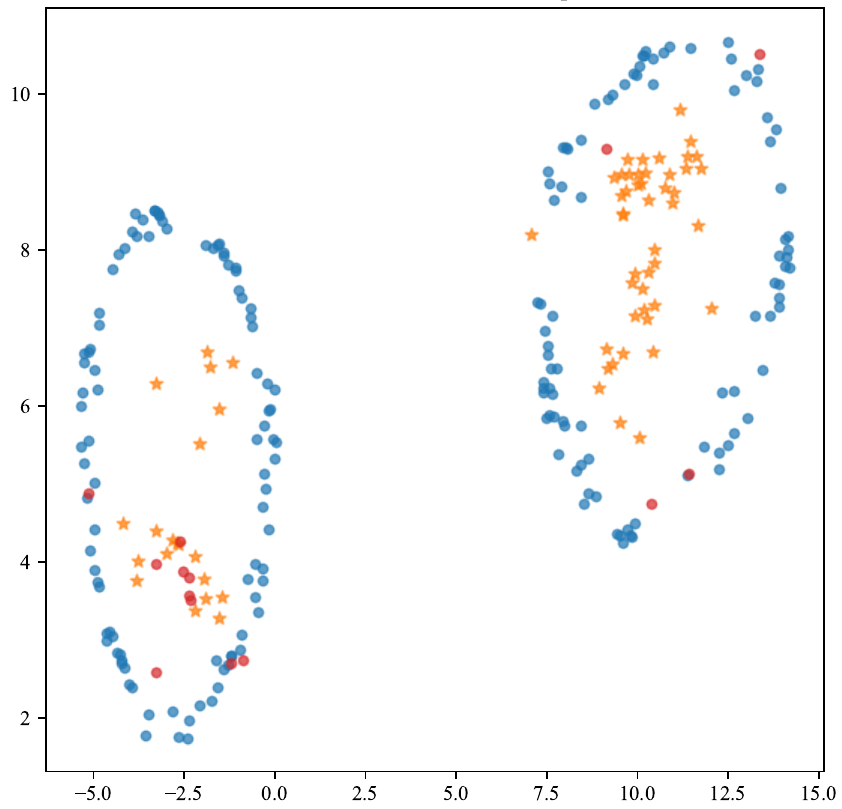}
            \label{fig:umap_capmix_esa}
        \end{minipage}%
    }
    \vspace{-0.4em}
    \caption{UMAP visualization of samples generated by \ourmethod (orange), along with normal (blue) and real anomalous (red) test samples. \ourmethod aligns better with real anomalies while largely avoiding overlap with normal regions.}
    \label{umap_pic}
\end{figure}

Overall, \ourmethod attains the best performance across datasets, reflecting the complementarity of label revision and DS-mixup: calibrated supervision improves anomaly plausibility, while dual-space interpolation increases sample diversity and generalization.

\subsubsection{Alleviation of Anomaly Shift.}

To further analyze whether \ourmethod can alleviate anomaly shift, we visualize UMAP embeddings of synthetic samples in Fig.~\ref{umap_pic}. The orange stars denote synthetic anomalies generated by \ourmethod, and the blue/red dots correspond to normal/real anomalous test samples.
We observe that \ourmethod produces pseudo-anomalies that align better with the real abnormal distribution while largely avoiding overlap with normal regions, which helps reduce the risk of learning a biased decision boundary. Especially in Fig.~\ref{umap_pic}(c), the orange points not only cover anomalous samples close to the normal cluster but also effectively inject anomalies in the lower-right area far away from the normal distribution, suggesting strong ability against anomaly shift.

\subsection{Robustness Analysis (RQ3)}
To further evaluate robustness under realistic conditions, we conduct a robustness study on the AIOps dataset, where training data naturally contains anomalous samples. We simulate different levels of data contamination by progressively injecting additional anomaly samples into the training set, thereby controlling the severity of training-set pollution.

\begin{figure}[t]
    \centering
    \subfigure[F1-score]{
    \begin{minipage}[t]{0.48\linewidth}
    \centering
    \includegraphics[width=\linewidth]
    {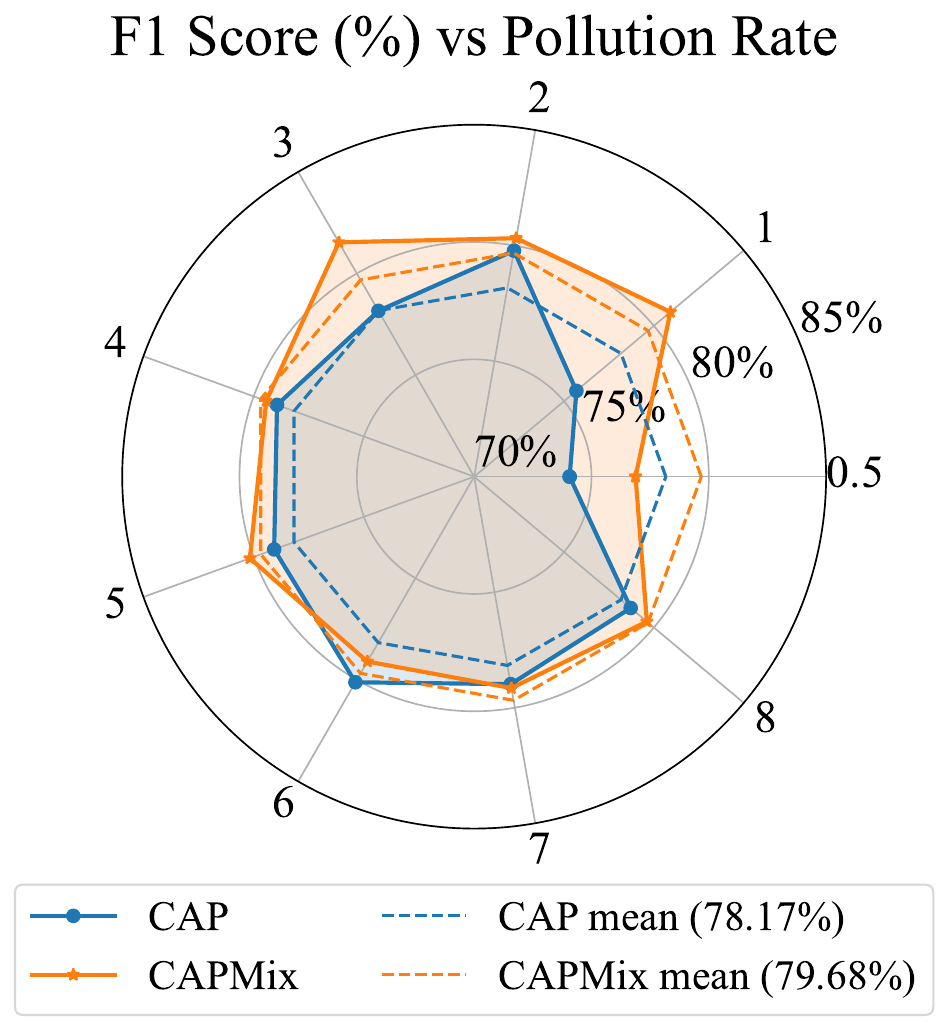}
        \vspace{-4pt}
    \label{fig:robust_f1}
    \end{minipage}
    }%
    \subfigure[Precision]{
    \begin{minipage}[t]{0.48\linewidth}
    \centering
    
    \includegraphics[width=\linewidth]
    {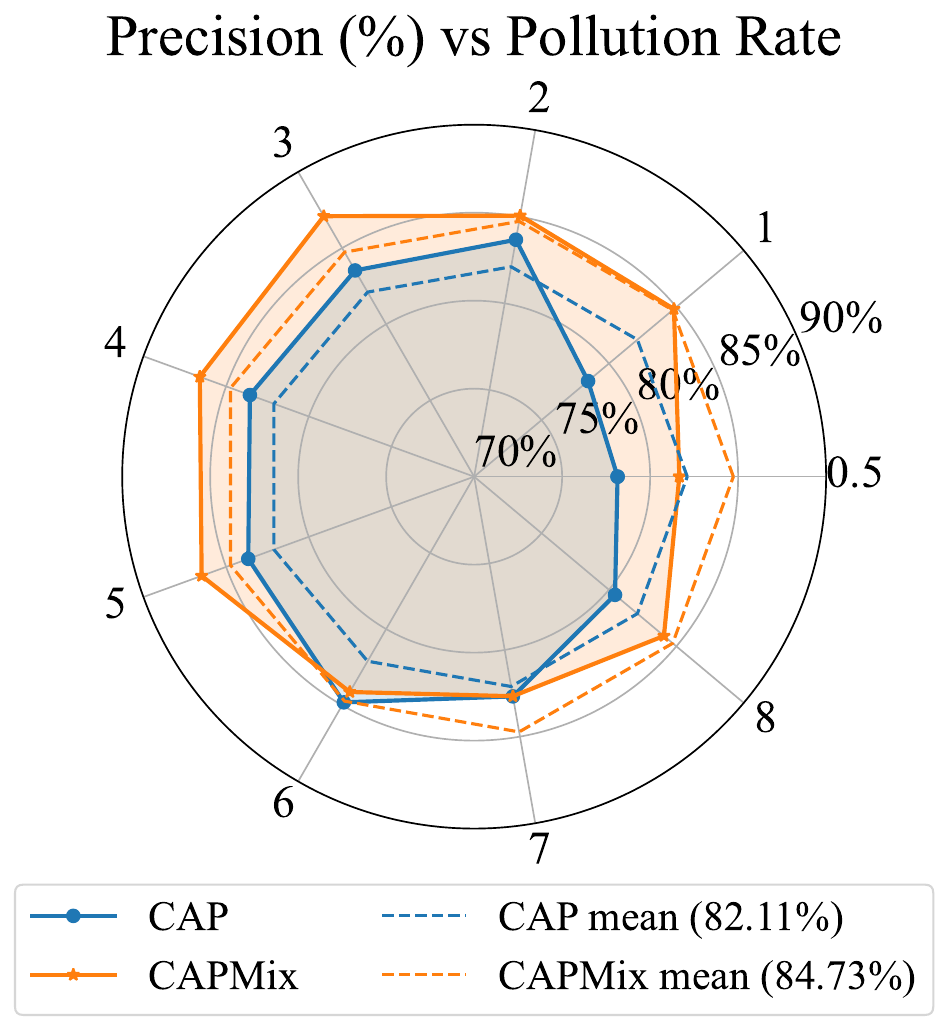}
        \vspace{-4pt}
    \label{fig:robust_pre}
    \end{minipage}
    }%
    \vspace{-0.4em}
    \caption{Robustness of \ourmethod vs. CutAddPaste (CAP) under increasing training-set contamination (0.5$\times$--8$\times$), evaluated by F1 and Precision.}
    \label{fig:robust}
\end{figure} 

As Fig.\ref{fig:robust} shows, both \ourmethod and the original CAP baseline demonstrate a certain degree of robustness under mild contamination. However, a clear performance gap emerges as the pollution becomes more severe. While CAP maintains relatively stable performance at lower corruption levels, its performance gradually degrades as the contamination increases. In contrast, \ourmethod consistently preserves higher F1-scores and precision across all settings, with the advantage becoming more pronounced at high pollution levels (e.g., 8×).

This trend suggests that \ourmethod is more resilient to heavily contaminated or drifted training distributions. The improvement can be attributed to two factors: (i) label revision mitigates the impact of noisy or misleading supervision, and (ii) dual-space mixup smooths the decision boundary by interpolating both input and representation spaces, reducing overfitting to corrupted samples.

Overall, these results indicate that \ourmethod not only inherits the robustness of CAP under moderate noise, but also provides additional stability under severe contamination, making it more suitable for real-world deployments where training data is often imperfect and continuously evolving.

\begin{figure}[t]
    \centering
    \subfigure[$\alpha$ and $\gamma$ on AIOps]{
    \begin{minipage}[t]{0.47\linewidth}
    \centering
    \includegraphics[width=\linewidth]
    {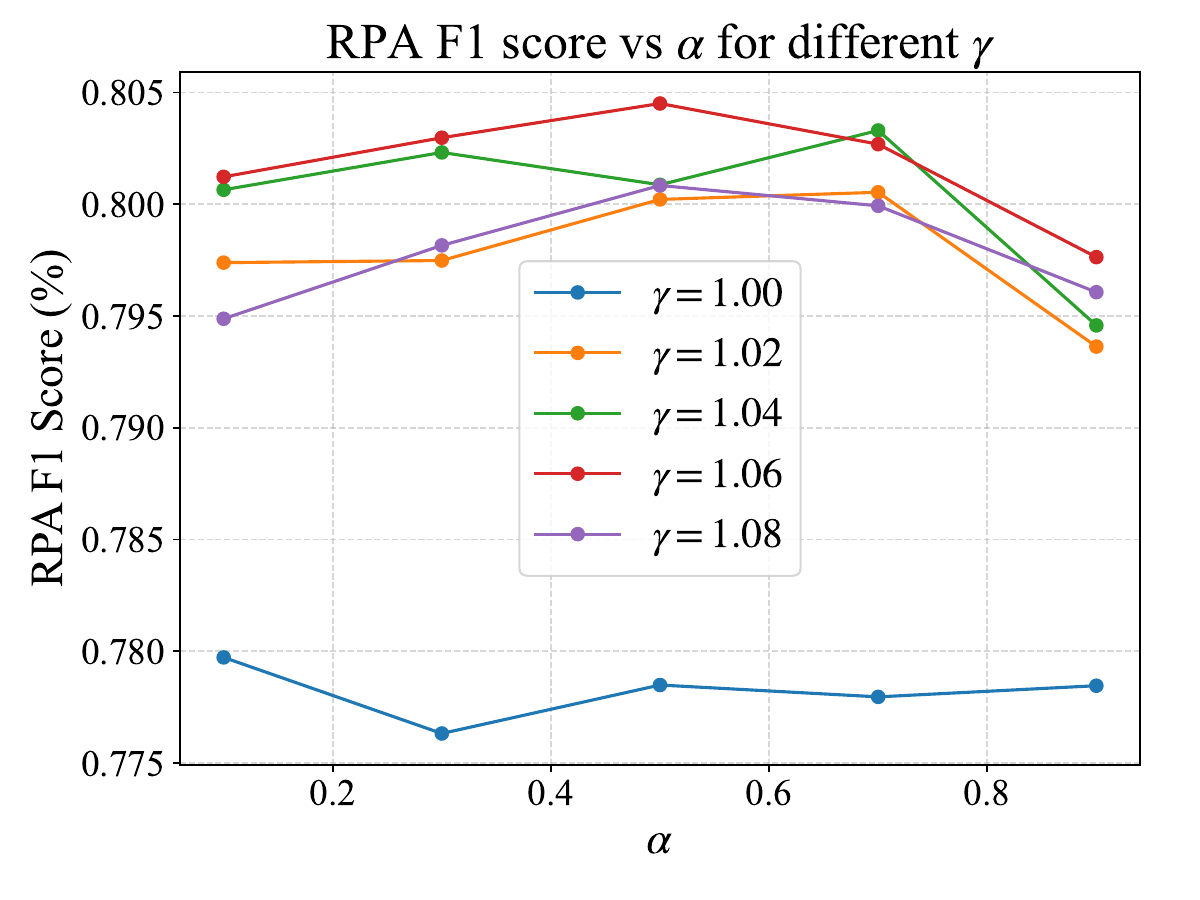}
    \label{fig:sense_kpi}
    \end{minipage}
    }
    \subfigure[$\alpha$ and $\gamma$ on WADI]{
    \begin{minipage}[t]{0.47\linewidth}
    \centering
    
    
    \includegraphics[width=\linewidth]
    {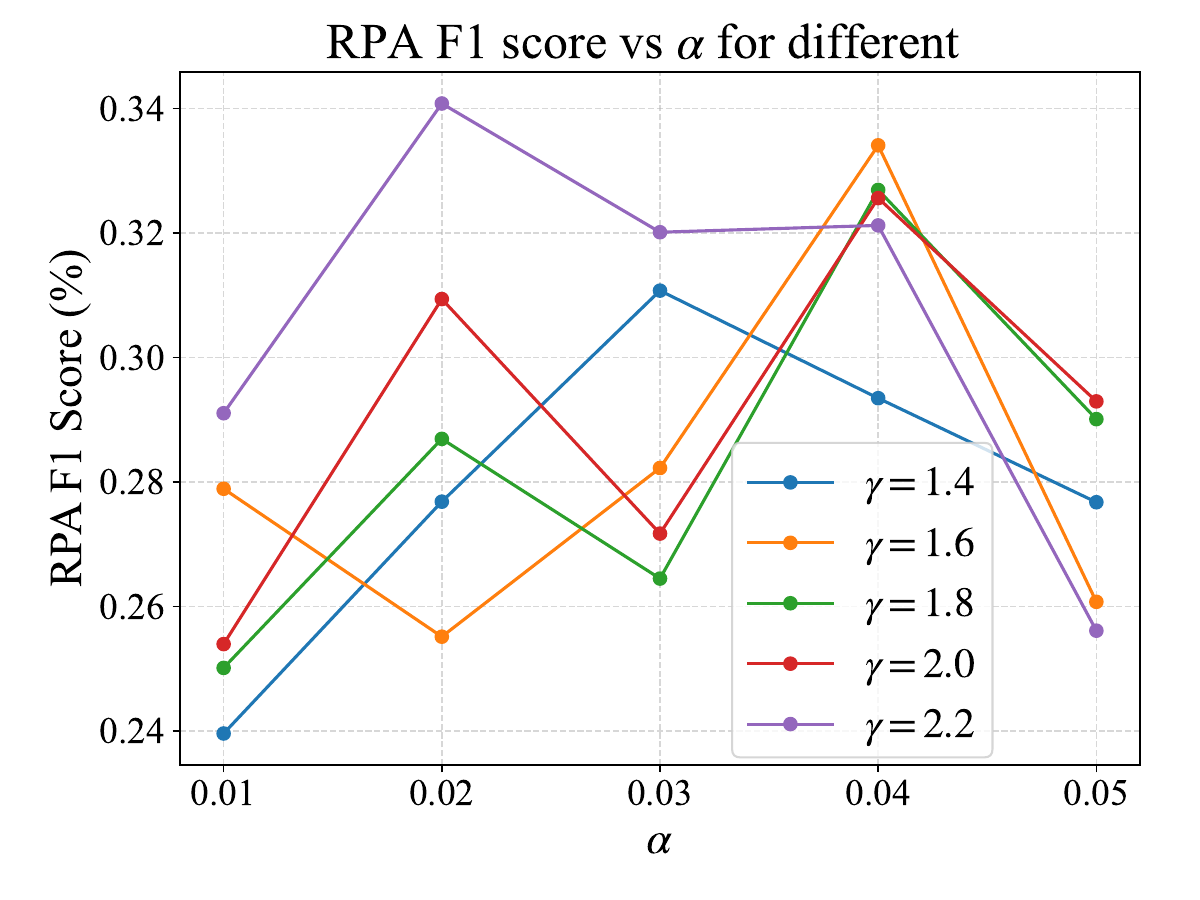}
    \label{fig:sense_wadi}
    \end{minipage}
    }%
    \vspace{-0.5em}
    \caption{Effectiveness of different $\gamma$ values (five lines) and $\alpha$ settings on the RPA F1 score.
Each line represents one $\gamma$; points along each line correspond to different $\alpha$ values.}
    \label{hyper_parameters}
    \vspace{-1em}
\end{figure}
\subsection{Hyperparameter Analysis (RQ4)}
We conduct a sensitivity analysis on two representative datasets, AIOps and WADI, to evaluate the impact of the label revision range ($\gamma$) and the mixup ratio ($\alpha$). These datasets exhibit distinct characteristics in terms of dimensionality and temporal complexity, enabling us to examine how the two components behave under different data regimes. For each dataset, we vary $\gamma$ across multiple levels and sweep $\alpha$ within a practical range; the results are shown in Fig.~\ref{hyper_parameters}. Other hyperparameters, such as the trend degree $\rho$ and minimum patch length $\zeta$, follow the settings in our previous work.

On AIOps, the performance curves under different $\gamma$ are clearly separated, indicating that the revision range is the dominant factor. In contrast, the model is relatively insensitive to $\alpha$, suggesting that mixup provides limited benefit when anomalies are simple and locally distinguishable.
On WADI, the trend differs: performance varies more noticeably with $\alpha$, highlighting the importance of dual-space mixup in capturing complex inter-variable dependencies. Meanwhile, $\gamma$ still provides complementary gains by constraining near-normal synthetic anomalies and stabilizing supervision.

Overall, these results indicate that the effectiveness of the two components depends on data characteristics. The revision range $\gamma$ primarily calibrates supervision for ambiguous or normal-like synthetic samples, while the mixup ratio $\alpha$ improves diversity and robustness, particularly in settings with complex temporal dependencies. The two components, therefore, play complementary roles across different data regimes.

\section{Case Study in Kuaishou}
\begin{figure}[t]
\centering
\includegraphics[width=0.95\linewidth]{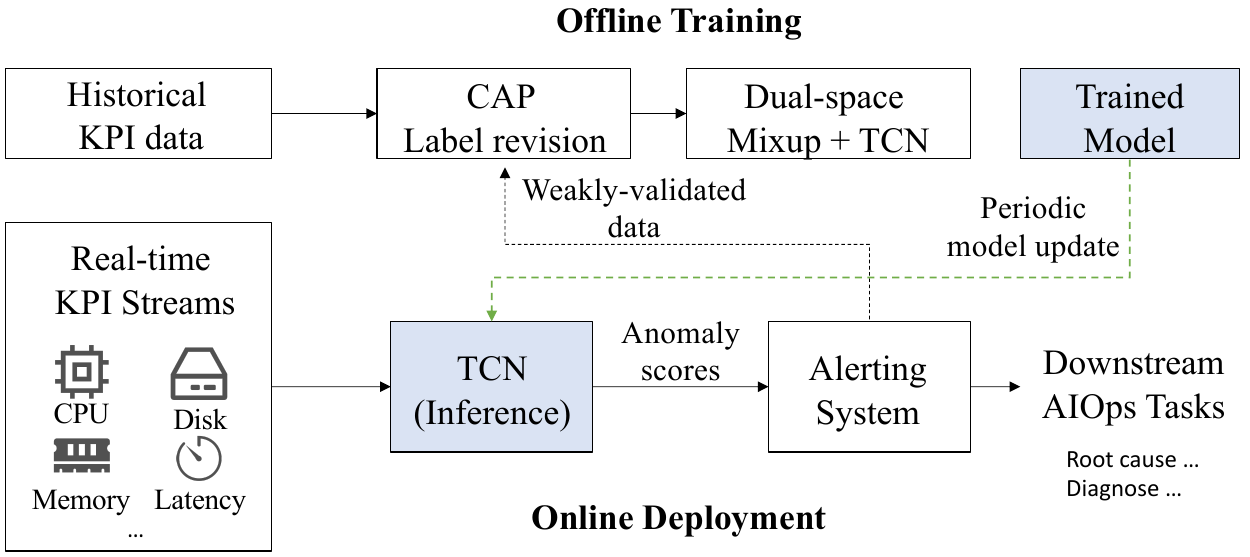}
\vspace{-0.4em}
\caption{Deployment overview of \ourmethod in KAIOps.}
\label{fig:deploy_flow}
\end{figure}

\ourmethod is integrated with KAIOps~\cite{wang2025kaiops}, the production-grade AIOps framework for Kuaishou’s AI clusters that accommodate large-scale model training jobs and real-time recommendation services. Figure~\ref{fig:deploy_flow} showcases the end-to-end workflow that encompasses both offline and online phases. In particular, a feedback loop is formed by alert validation results, such as operator-confirmed false positives and true incidents, which are periodically aggregated and used to refine the label revision stage. In this section, we provide a qualitative case analysis and a controlled evaluation dataset to demonstrate robustness under real operational fluctuations.

\subsection{Applying \ourmethod in  Large-scale AI Clusters}

\subsubsection{KPI Streams in AI Clusters} 

We collected KPI streams generated by production-grade workloads, together with resource utilization logs, from industrial-scale AI computing clusters. 

The measurements are inherently noisy, and the system dynamics exhibit temporal drift. Since most events correspond to benign operational behaviors, the setting is highly contaminated, i.e., normal events often exhibit anomaly-like patterns. Therefore, we conduct chaos engineering by intentionally introducing failures to validate the infrastructure's resilience.

We injected anomalies by executing representative fault workloads directly on the target nodes.  This approach preserves natural system dynamics and yields KPI traces that reflect realistic fault behaviors. Because faults are injected at the node level, anomalies often affect only a subset of metrics while others remain normal, resulting in cross-metric inconsistency and partial observability. Combined with the prevalence of benign fluctuations, this creates a challenging evaluation scenario with noisy, ambiguous, and weakly structured anomalies.

To ensure reproducibility, we publicly released a dataset covering a continuous period of 16 days with a temporal resolution of one minute.  A statistical overview of the dataset is provided in Table~\ref{tab:dataset_stats}. The dataset consists of node-level, multivariate KPI traces from a representative and heterogeneous subset of seven nodes. These KPIs include, for example, CPU utilization, memory consumption, disk input/output (I/O) metrics, and network traffic measurements. In addition, the dataset contains cluster-level observability indicators, such as the number of pending tasks, API request latency, and the number of container restart events.

\subsubsection{\ourmethod's Performance}  We evaluate \ourmethod's performance in tackling real-world KPI data streams. The streams encompass both genuine anomalous events and regular operational events (such as metric scraping and log rotation) that frequently induce transient perturbations in the observed metrics, which are close to failure patterns.

\begin{table}[t]
\caption{Statistical overview of the AIClusterKPI dataset.}
\label{tab:dataset_stats}
\centering
\scalebox{0.9}{\begin{tabular}{cc}
\toprule
\multicolumn{2}{c}{\textit{\textbf{Dataset Configuration }}} \\
\midrule
\textbf{Total Data} & 20,452 points \\
\textbf{Training Set} & 17,571 points  \\
\textbf{Test Set} & 3,029 points / 598 events\\
\textbf{Test Anomaly} & 1,117 points / 145 events \\ \midrule
\textbf{Node Signals} & CPU, Memory, I\/O, Network \\
\textbf{Cluster Signals} & API Latency, Pending Tasks, Pod Restarts \\
\textbf{Label Granularity} & Binary (0/1), Fault Type, Node ID \\ 
\textbf{Typical Faults} & CPU\_high, Mem\_leak, IO\_burst \\
\midrule
\multicolumn{2}{c}{\textit{\textbf{\ourmethod Performance as Baseline}}} \\
\midrule
\textbf{Detection Accuracy} & RPA Precision: 0.80 / F1-score: 0.79\\
\textbf{Alert Reliability} & PR-AUC: 0.87 / NAB Score: 0.76\\
\bottomrule
\end{tabular}}
\vspace{-1.4em}
\end{table}

\begin{figure}[t]
    \centering
\includegraphics[width=1\columnwidth]{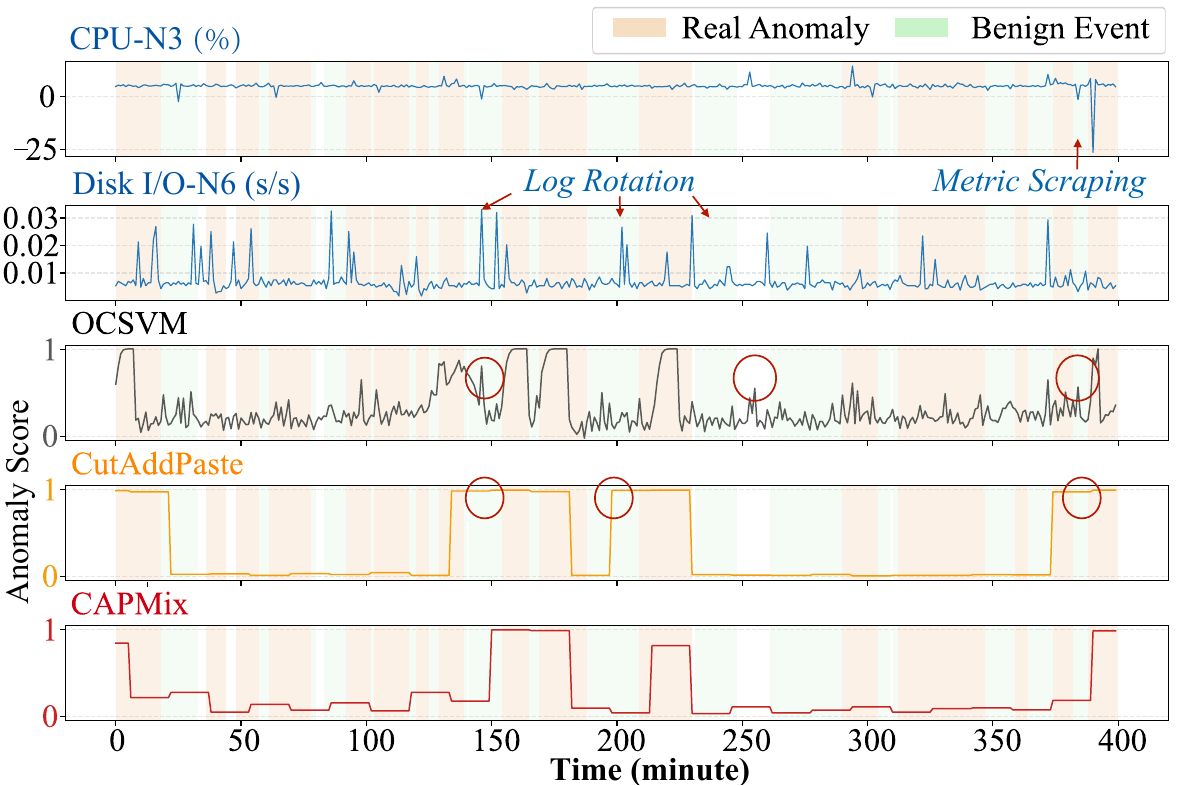}
    \caption{Detection performance on real-world data. \ourmethod successfully suppresses false alarms during benign events.}
    \label{visual}
\vspace{-1.6em}
\end{figure}

As shown in Fig.~\ref{visual}, the first two rows depict raw KPI signals from two nodes with annotated intervals. Dark-shaded regions indicate anomalous periods, such as those induced by memory leakage, whereas light-shaded regions correspond to benign events. Notably, certain extreme spikes (e.g., negative CPU values) arise from measurement noise or monitoring artifacts, rather than actual system events. The last three rows present a comparison of anomaly scores produced by \ourmethod against scores by a representative point-wise baseline, OC-SVM, and CutAddPaste.

OC-SVM exhibits frequent oscillations in its anomaly scores across both anomalous and non-anomalous regions, leading to spurious peaks during benign events. CutAddPaste baseline partially mitigates this phenomenon but still produces undesirably high scores in several non-anomalous intervals. In contrast, \ourmethod produces consistently low scores in the presence of transient noise-like fluctuations while producing clear, temporally sustained responses to genuine anomalous events.
Around time point 250, a scheduled log rotation triggers a sharp, but operationally benign, spike in disk I/O. OC-SVM responds with pronounced score fluctuations, whereas \ourmethod remains comparatively stable, indicating enhanced robustness to such benign perturbations. A similar pattern is observed in the benign intervals preceding indices 150 and 200, where both OC-SVM and CutAddPaste assign elevated anomaly scores. Likewise, near index 400, metric scraping again introduces irregular peaks, resulting in a high anomaly score for CutAddPaste. \ourmethod maintains a lower and more stable response.

We report raw outputs without post-processing, highlighting the intrinsic behavior of each method. Overall, \ourmethod demonstrates improved robustness to operational noise and more reliable alignment with sustained anomaly intervals, indicating its potential to reduce false alarms in real-world deployments.

\subsection{Engineering Lessons and Experiences}
We summarize several lessons from deploying \ourmethod\ in Kuaishou.

\textit{Robust boundary learning is more effective than clean-data assumptions.}
Production KPI streams inevitably contain noise and transient fluctuations, under which methods relying on clean training data tend to overreact and trigger false alarms. We observe that learning robust decision boundaries under contamination is more effective in practice. By combining CAP-based augmentation with Dual-Space Mixup, \ourmethod\ focuses on structural failure patterns while remaining insensitive to benign fluctuations, which is consistent with the reduction in false positives observed above.

\textit{Labels should be treated as uncertain supervision signals.}
Industrial anomaly labels are sparse, delayed, and often unreliable, and pseudo-anomalies may overlap with real failures. Treating such labels as ground truth can degrade performance. The Label Revision (LR) mechanism addresses this issue by refining supervision based on representation consistency, improving stability under noisy and evolving environments.

\textit{Computational efficiency is a primary constraint in deployment.}
In SLA-critical monitoring systems, inference cost directly affects usability at scale. We observe that heavyweight architectures can introduce significant latency under peak workloads. By adopting a lightweight TCN backbone with efficient dual-space operations, \ourmethod\ achieves low-latency inference, enabling practical deployment in large-scale AI clusters.

\section{Conclusions}\label{section conclusion}
This paper addresses the challenge of robust KPI anomaly detection in noisy and non-stationary AIOps environments. By formalizing \textit{Anomaly Shift}, we highlight the critical gap between synthetic training patterns and evolving real-world failures. To bridge this gap, we propose \ourmethod, a controllable augmentation framework that integrates prior-guided injection, DTW-based label refinement, and Dual-Space Mixup into a lightweight TCN architecture. Extensive evaluations across eight public datasets demonstrate that \ourmethod\ significantly outperforms state-of-the-art baselines in Best RPA-F1 score and maintains exceptional resilience under data contamination. Beyond academic metrics, \ourmethod\ has been successfully integrated into Kuaishou's production AI clusters, where it effectively mitigated alert fatigue and reclaimed operational trust. By releasing the AIClusterKPI dataset and our codebase, we aim to provide the community with a more realistic dataset for advancing robust and deployment-ready anomaly detection.
\section{Mandatory Data Availability Statement}

AIClusterKPI is a dataset derived from industrial production settings, containing 16 days of multi-dimensional metrics with real dynamic system changes.
Sensitive metadata is de-identified, and metrics are normalized; the code is publicly available on GitHub (\url{https://github.com/alsike22/CAPMix}) and the data is archived on Zenodo (\url{https://doi.org/10.5281/zenodo.19926114}). All datasets used in our main experiments (in Table~\ref{results}) are public benchmarks.

\section*{Acknowledgment} 

We would very much like to thank anonymous reviewers for their valuable  comments. This work is supported in part by National Key
R\&D Program of China (Grant No. 2024YFB4505901), in part by NSFC (Grant No. 62402024, 62302485), in part by the Fundamental Research Funds for the Central Universities, in part by the Key Research Project of Chinese Academy of Sciences (No. RCJJ-145-24-21), and, last but not the least, by Kuaishou Research Fund. For any correspondence, please refer to the project lead and coordinator Prof. Renyu Yang (renyuyang@buaa.edu.cn).



\bibliographystyle{unsrt}
\balance
\bibliography{ref}



\end{document}